\documentclass[a4paper,fleqn]{cas-dc}
\usepackage{microtype}
\usepackage[numbers, square]{natbib}  %
\usepackage{graphicx}  %
\usepackage{changepage}
\usepackage{subfiles}
\usepackage{IEEEtrantools}
\usepackage{tikz}
\usepackage[edges]{forest}
\usepackage{float}     %
\usepackage{xcolor}
\usepackage{xcolor}%
\usepackage{textcomp}%
\usepackage{manyfoot}%
\usepackage{booktabs}%
\usepackage{algorithm}%
\usepackage{algorithmicx}%
\usepackage{algpseudocode}%
\usepackage{listings}%
\usepackage{changepage}
\usepackage{threeparttable}
\usepackage{colortbl}
\usepackage{subfiles}
\usepackage{IEEEtrantools}
\usepackage{tikz}
\usepackage{url}
\usepackage[edges]{forest}
\usepackage{float}     %
\usepackage{xcolor}
\usepackage{dashrule}
\usepackage{tcolorbox}
\def\tsc#1{\csdef{#1}{\textsc{\lowercase{#1}}\xspace}}
\tsc{WGM}
\tsc{QE}
\tsc{EP}
\tsc{PMS}
\tsc{BEC}
\tsc{DE}
\begin{document}
\let\WriteBookmarks\relax
\def\floatpagepagefraction{1}
\def\textpagefraction{.001}

\shorttitle{A Survey on Image-text Multimodal Models}

\title [mode = title]{A Survey on Advancements in Image-Text Multimodal Models: From General Techniques to Biomedical Implementations 
}                      
\author[inst1,inst2]{Ruifeng Guo}
\ead{grf@sict.ac.cn}
\cortext[corcorr]{Corresponding author.}
\author[inst1,inst2]{Jingxuan Wei}
\ead{weijingxuan20@mails.ucas.edu.cn}
\cormark[1]

\author[inst1,inst2]{Linzhuang Sun}
\ead{sunlinzhuang21@mails.ucas.ac.cn}

\author[inst1,inst2]{Bihui Yu}
\ead{yubihui@sict.ac.cn}

\author[inst1,inst2]{Guiyong Chang}
\ead{changguiyong22@mails.ucas.ac.cn}

\author[inst1,inst2]{Dawei Liu}
\ead{liudawei22@mails.ucas.ac.cn}

\author[inst1,inst2]{Sibo Zhang}
\ead{zhangsibo22@mails.ucas.ac.cn}

\author[inst1,inst2]{Zhengbing Yao}
\ead{yaozhengbing22@mails.ucas.ac.cn}

\author[inst1,inst2]{Mingjun Xu}
\ead{xumingjun22@mails.ucas.ac.cn}

\author[inst1,inst2]{Liping Bu}
\ead{buliping@sict.ac.cn}

\nonumnote{All authors contributed equally to this work.}
\affiliation[inst1]{organization={Shenyang Institute of Computing Technology, Chinese Academy of Sciences},
    city={Shenyang},
    postcode={110168}, 
    country={China}}

\affiliation[inst2]{organization={University of Chinese Academy of Sciences},
    city={Beijing},
    postcode={100049}, 
    country={China}}

\begin{abstract}
With the significant advancements of Large Language Models (LLMs) in the field of Natural Language Processing (NLP), the development of image-text multimodal models has garnered widespread attention. Current surveys on image-text multimodal models mainly focus on representative models or application domains, but lack a review on how general technical models influence the development of domain-specific models, which is crucial for domain researchers. Based on this, this paper first reviews the technological evolution of image-text multimodal models, from early explorations of feature space to visual language encoding structures, and then to the latest large model architectures. Next, from the perspective of technological evolution, we explain how the development of general image-text multimodal technologies promotes the progress of multimodal technologies in the biomedical field, as well as the importance and complexity of specific datasets in the biomedical domain. Then, centered on the tasks of image-text multimodal models, we analyze their common components and challenges. After that, we summarize the architecture, components, and data of general image-text multimodal models, and introduce the applications and improvements of image-text multimodal models in the biomedical field. Finally, we categorize the challenges faced in the development and application of general models into external factors and intrinsic factors, further refining them into 2 external factors and 5 intrinsic factors, and propose targeted solutions, providing guidance for future research directions. For more details and data, please visit our GitHub page: \url{https://github.com/i2vec/A-survey-on-image-text-multimodal-models}.
\end{abstract}


\maketitle

\section{Introduction}

In recent years, large language models (LLMs)~\cite{zhao2023survey,min2023recent,chang2023survey,kasneci2023chatgpt} such as ChatGPT~\cite{brown2020language}, ChatGLM~\cite{zeng2022glm}, and LLaMA~\cite{touvron2023llama} have made impressive and rapid progress, and are highly regarded for their performance in natural language processing (NLP)~\cite{patil2023survey,madsen2022post,zhang2023survey,shao2022tracing} tasks. While LLM has demonstrated impressive capabilities, it has notable limitations when it comes to processing and responding to image data. To address this phenomenon, researchers have developed various image-text multimodal models. A multimodal model is an artificial intelligence model that is capable of processing multiple types of data. It fuses and associates data from different perceptual modalities (e.g., visual, verbal, audio, etc.) in order to achieve more comprehensive and accurate information understanding and reasoning capabilities. At present, image-text  multimodal research represents the principal focus of multimodal research~\cite{cao2022image,ying2022survey,zhang2024vision}. 

The development of image-text multimodal models has garnered significant attention ~\cite{li2023multimodal}. Initially, to explore the joint feature space of learned images and their descriptions,  kernel canonical correlation analysis (KCCA)~\cite{hodosh2013framing} was used to project image and sentence features into a common space by learning projections. Subsequent research has attempted bi-directional mapping between images and text descriptions using CNN and RNN~\cite{chen2015mind,mao2014explain,karpathy2015deep}. However, it was found that RNN had weaknesses in remembering concepts. Therefore, Long Short-Term Memory Networks (LSTM) were introduced to form CNN and LSTM~\cite{venugopalan2014translating}. The introduction of the attention~\cite{vaswani2017attention} mechanism has significantly improved the capabilities of image-text multimodal models. Graphical descriptions based on CNN and LSTM can now be trained by maximizing an approximate variational lower bound or, equivalently, by reinforcement~\cite{xu2015show}. The emergence of the Transformer~\cite{lu2021transformer} model was a significant milestone in the development of image-text multimodal modeling. The Transformer dramatically improved the performance of image-text multimodal models, particularly in language understanding. Models such as Bert~\cite{lan2019albert} have revolutionized the field of natural language processing and have subsequently been applied to multimodal scenarios, such as ViLBERT (Visual and Language BERT)~\cite{lu2019vilbert,sarzynska2021detecting}. These models can process complex linguistic data and understand image content, which can be applied to advanced visual quizzes and image-text matching tasks. Image-text multimodal models is now entering the era of large models that rely on two key components: visual encoder~\cite{dosovitskiy2020image,li2023blip,badrinarayanan2017segnet} and LLM~\cite{zeng2022glm,touvron2023llama,chowdhery2023palm}. The LLM is responsible for processing natural language text, while the visual coder module extracts and interprets visual information from images. 
BLIP-2, as a pioneering work, successfully connects visual models to large language models by introducing a lightweight query-former (Q-former). InstructBLIP proposes a new instruction fine-tuning paradigm based on BLIP-2, which demonstrates the extraction of more useful visual features by utilizing additional instructions. Meanwhile, LLaVA attempts to use more concise connectors, such as linear layers or MLPs, to link visual models to large language models, again with equally remarkable results.

\definecolor{hidden-draw}{RGB}{0,0,255} 
\definecolor{hidden-pink}{RGB}{1.0, 0.41, 0.71}
\tikzstyle{my-box}=[
    rectangle,
    rounded corners,
    text opacity=1,
    minimum height=1.5em,
    minimum width=5em,
    inner sep=2pt,
    align=center,
    fill opacity=.5,
    line width=0.8pt,
]
\tikzstyle{leaf}=[my-box, minimum height=1.5em,
    fill=hidden-pink!80, text=black, align=left,font=\normalsize,
    inner xsep=2pt,
    inner ysep=4pt,
    line width=0.8pt,
]
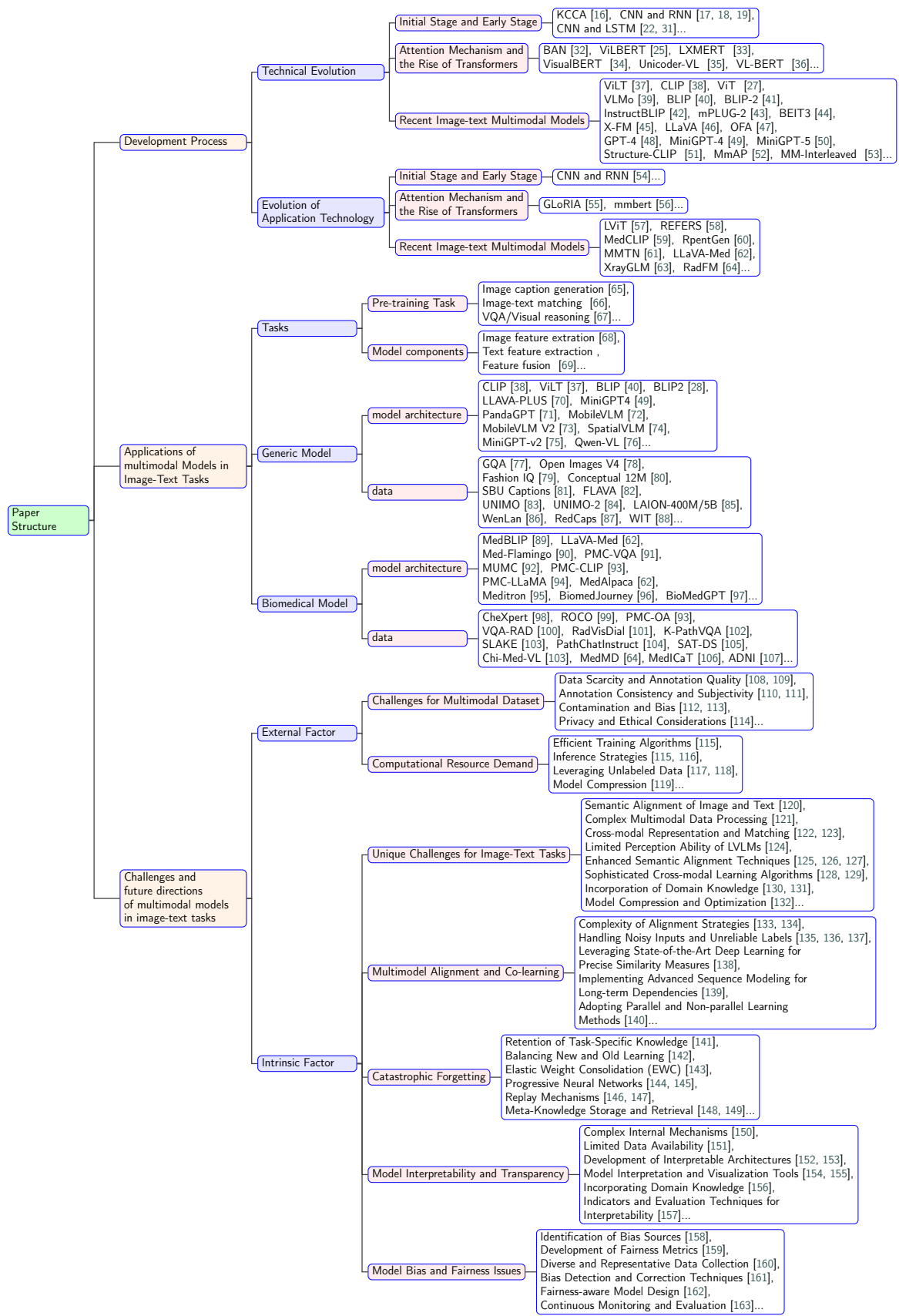
\begin{figure*}[!th]
     \begin{adjustwidth}{2em}{-2em}  
        \resizebox{0.88\textwidth}{!}{  
        \begin{forest}
            forked edges,
            for tree={
                grow=east,
                reversed=true,
                anchor=west,
                parent anchor=east,
                child anchor=west,
                base=left,
                font=\large,
                rectangle,
                draw=hidden-draw,
                rounded corners,
                align=left,
                minimum width=4em,
                edge+={darkgray, line width=1pt},
                s sep=5pt,
                inner xsep=4pt,
                inner ysep=1pt,
                line width=0.8pt,
                ver/.style={rotate=90, child anchor=north, parent anchor=south, anchor=center},
            },
            where level=1{text width=5em,fill=orange!10}{},
            where level=2{text width=5em,fill=blue!10}{},
            where level=3{yshift=0.26pt,fill=pink!30}{},
            where level=4{yshift=0.26pt,fill=yellow!20}{},
            where level=5{yshift=0.26pt}{},
            [
               Paper \\
               Structure, fill=green!20,text width=8em
                [
                    Development Process, text width=13em,l=4cm
                    [
                        Technical Evolution, text width=13em,l=5em
                        [
                            Initial Stage and Early Stage
                            [
                            KCCA~\cite{hodosh2013framing}{, }
                            CNN and RNN~\cite{chen2015mind,mao2014explain,karpathy2015deep}{, }\\
                            CNN and LSTM~\cite{xu2015show,vinyals2015show}{... } \\
                            ,fill=green!0
                            ] 
                        ]
                        [
                        Attention Mechanism and \\
                        the Rise of Transformers
                            [
                            BAN~\cite{kim2018bilinear}{, } ViLBERT~\cite{lu2019vilbert}{, }
                            LXMERT ~\cite{tan2019lxmert}{, }\\
                            VisualBERT ~\cite{li2019visualbert}{, } 
                            Unicoder-VL ~\cite{li2020unicoder}{, } VL-BERT ~\cite{DBLP:conf/iclr/SuZCLLWD20}{... }
                            ,fill=green!0
                            ]
                        ]
                        [
                        Recent Image-text Multimodal Models
                            [ 
                            ViLT~\cite{kim2021vilt}{, }
                            CLIP~\cite{radford2021learning}{, } ViT ~\cite{dosovitskiy2020image}{, }\\
                            VLMo~\cite{bao2022vlmo}{, }
                            BLIP~\cite{Li_Li_Xiong_Hoi}{, }
                            BLIP-2~\cite{Li_Li_Savarese_Hoi}{, }\\
                            InstructBLIP~\cite{dai2305instructblip}{, }
                            mPLUG-2~\cite{xu2023mplug}{, }
                            BEIT3~\cite{wang2022image}{, }\\
                            X-FM~\cite{chen2022learning}{, }
                            LLaVA~\cite{Liu_Li_Wu_Lee_-Madison_Research}{, }
                            OFA~\cite{wang2022ofa}{, }\\
                            GPT-4~\cite{bubeck2023sparks}{, }
                            MiniGPT-4~\cite{zhu2023minigpt}{, }
                            MiniGPT-5~\cite{zheng2023minigpt}{, }\\
                            Structure-CLIP ~\cite{huang2023structure}{, }  
                            MmAP~\cite{xin2023mmap}{, } 
                            MM-Interleaved ~\cite{tian2024mm}{... }
                            ,fill=green!0
                            ]
                        ]
                    ]
                    [
                    Evolution of \\
                    Application Technology\\
                     ,text width=13em,l=5em
                        [
                        Initial Stage and Early Stage
                            [
                            CNN and RNN~\cite{gheisari2021combined}{... }
                            ,fill=green!0
                            ]
                        ]
                        [
                        Attention Mechanism and \\
                        the Rise of Transformers
                            [
                                GLoRIA~\cite{huang2021gloria}{, }
                                mmbert~\cite{khare2021mmbert}{...  }
                                ,fill=green!0
                            ]
                        ]
                        [
                        Recent Image-text Multimodal Models
                            [
                        LViT~\cite{li2023lvit}{, }
                        REFERS~\cite{zhou2022generalized}{, }\\
                        MedCLIP~\cite{wang2022medclip}{, }
                        RpentGen~\cite{chambon2022roentgen}{, }\\
                        MMTN~\cite{cao2023mmtn}{, }
                        LLaVA-Med~\cite{li2023llava}{, }\\
                        XrayGLM~\cite{wang2023XrayGLM}{, }
                        RadFM~\cite{wu2023towards}{... }
                        ,fill=green!0
                        ]
                        ]
                    ]
                ]
            [                
                    Applications of \\
                    multimodal Models in \\ 
                    Image-Text Tasks, text width=13em,l=4cm
                    [
                        Tasks , text width=10em
                        [
                            Pre-training Task, text width=10em
                            [
                           Image caption generation~\cite{farhadi2010every}{, }\\
                           Image-text matching ~\cite{diao2021similarity}{, }\\
                           VQA/Visual reasoning~\cite{cheng2022visual}{... }
                            ,fill=green!0
                            ]
                        ]
                        [
                            Model components, text width=10em
                            [
                            Image feature extration~\cite{simonyan2014very}{, }\\
                            Text feature extraction {, }\\
                            Feature fusion ~\cite{zhu2020deformable}{... }
                            ,fill=green!0
                            ]
                        ]
                    ]
                    [
                        Generic Model , text width=10em
                        [
                           model architecture, text width=10em
                            [
                            CLIP~\cite{radford2021learning}{, }
                            ViLT~\cite{kim2021vilt}{, }
                              BLIP~\cite{Li_Li_Xiong_Hoi}{, } 
                              BLIP2~\cite{li2023blip}{, } \\
                              LLAVA-PLUS~\cite{liu2023llava}{, } 
                              MiniGPT4~\cite{zhu2023minigpt}{, }\\
                              PandaGPT~\cite{su2023pandagpt}{, } 
                              MobileVLM~\cite{chu2023mobilevlm}{, } \\
                              MobileVLM V2~\cite{chu2024mobilevlm}{, }                             SpatialVLM~\cite{chen2024spatialvlm}{, } \\
                              MiniGPT-v2~\cite{chen2023minigpt}{, }
                              Qwen-VL~\cite{bai2023qwen}{... } \\
                            ,fill=green!0
                            ]
                        ]
                        [
                            data, text width=10em
                            [
                            GQA~\cite{hudson2019gqa}{, } 
                            Open Images V4~\cite{kuznetsova2020open}{, } \\
                            Fashion IQ~\cite{wu2021fashion}{, }
                            Conceptual 12M~\cite{changpinyo2021conceptual}{, } \\
                            SBU Captions~\cite{ordonez2011im2text}{, } 
                            FLAVA~\cite{singh2022flava}{, } \\
                            UNIMO~\cite{li2020unimo}{, }
                            UNIMO-2~\cite{li2022unimo}{, } 
                            LAION-400M/5B~\cite{schuhmann2022laion}{, } \\
                            WenLan~\cite{huo2021wenlan}{, } 
                            RedCaps~\cite{desai2021redcaps}{, }
                            WIT~\cite{srinivasan2021wit}{... } 
                            ,fill=green!0
                            ]
                        ]
                    ]
                [
                        Biomedical Model , text width=10em
                        [
                           model architecture, text width=10em
                            [
                            MedBLIP~\cite{chen2023medblip}{, } 
                            LLaVA-Med~\cite{li2023llava}{, } \\
                            Med-Flamingo~\cite{moor2023med}{, } 
                            PMC-VQA~\cite{zhang2023pmc}{, } \\
                            MUMC~\cite{li2023masked}{, } 
                            PMC-CLIP~\cite{lin2023pmc}{, }\\
                            PMC-LLaMA~\cite{wu2023pmc}{, } 
                            MedAlpaca~\cite{li2023llava}{, } \\
                            Meditron~\cite{chen2023meditron}{, }
                            BiomedJourney~\cite{gu2023biomedjourney}{, } 
                            BioMedGPT~\cite{luo2023biomedgpt}{... }\\
                            ,fill=green!0
                            ]
                        ]
                        [
                             data, text width=10em
                            [
                            CheXpert~\cite{irvin2019chexpert}{, }
                            ROCO~\cite{pelka2018radiology}{, } 
                            PMC-OA~\cite{lin2023pmc}{, } \\
                            VQA-RAD~\cite{lau2018dataset}{, } 
                            RadVisDial~\cite{johnson2019mimic}{, }
                            K-PathVQA~\cite{naseem2023k}{, } \\
                            SLAKE~\cite{liu2021slake}{, } 
                    PathChatInstruct~\cite{lu2023foundational}{, }
                            SAT-DS~\cite{zhao2023one}{, } \\
                            Chi-Med-VL~\cite{liu2021slake}{, }
                            MedMD~\cite{wu2023towards}{,}
                            MedICaT~\cite{subramanian2020medicat}{,}
                            ADNI~\cite{petersen2010alzheimer}{...}
                            ,fill=green!0
                            ]
                        ]
                ]
            ]
            [
                  Challenges and \\
                  future directions \\
                  of multimodal models \\
                  in image-text tasks, text width=13em,l=4cm
                [
                    External Factor , text width=10em
                    [
                    Challenges for Multimodal Dataset
                        [
                        Data Scarcity and Annotation Quality~\cite{baltruvsaitis2018multimodal,bayoudh2022survey}{, }\\
                        Annotation Consistency and Subjectivity~\cite{miceli2020between,metallinou2013annotation}{, }\\
                        Contamination and Bias~\cite{xu2022algorithmic,jacoba2023bias}{, }\\
                        Privacy and Ethical Considerations~\cite{dhirani2023ethical}{... }\\
                        ,fill=green!0
                        ]
                    ]
                    [
                    Computational Resource Demand
                        [
                        Efficient Training Algorithms~\cite{li2023model}{,}\\
                        Inference Strategies~\cite{li2023model,zhu2023survey}{, }\\
                        Leveraging Unlabeled Data~\cite{ren2023weakly,taha2023semi}{, }\\
                        Model Compression~\cite{msuya2023deep}{... }\\
                        ,fill=green!0
                        ]
                    ]
                ]
                [
                    Intrinsic Factor, text width=10em
                    [
                        Unique Challenges for Image-Text Tasks
                        [
                        Semantic Alignment of Image and Text~\cite{yarom2024you}{, }\\
                        Complex Multimodal Data Processing~\cite{azad2023foundational}{, }\\
                        Cross-modal Representation and Matching~\cite{zhao2020cross,sun2023scoping}{, }\\
                        Limited Perception Ability of LVLMs~\cite{liu2024survey}{, }\\
                        Enhanced Semantic Alignment Techniques~\cite{choi2023transformer,he2023transformers,shamshad2023transformers}{, }\\
                        Sophisticated Cross-modal Learning Algorithms~\cite{bayoudh2023survey,karthikeyan2024novel}{, }\\
                        Incorporation of Domain Knowledge~\cite{cai2023incorporating,murali2023towards}{, }\\
                        Model Compression and Optimization~\cite{sung2023ecoflap}{... }\\
                        ,fill=green!0
                        ]
                    ]
                    [
                        Multimodel Alignment and Co-learning
                        [
                        Complexity of Alignment Strategies~\cite{wang2008aligning,shaik2023survey}{, }\\
                        Handling Noisy Inputs and Unreliable Labels~\cite{tao2019resilient,li2020label,nagarajan2024bayesian}{, }\\
                        Leveraging State-of-the-Art Deep Learning
                        for \\
                        Precise Similarity Measures~\cite{chen2023survey}{, }\\
                        Implementing Advanced Sequence Modeling for \\
                        Long-term Dependencies~\cite{tiezzi2024resurgence}{, }\\
                        Adopting Parallel and Non-parallel Learning \\
                        Methods~\cite{song2024multi}{... }\\
                        ,fill=green!0
                        ]
                    ]
                    [
                        Catastrophic Forgetting
                        [
                        Retention of Task-Specific Knowledge~\cite{mellal2023cnn}{, }\\
                        Balancing New and Old Learning~\cite{peng2021multiscale}{, }\\
                        Elastic Weight Consolidation (EWC)~\cite{yin2021mitigating}{, }\\
                        Progressive Neural Networks~\cite{sharma2023advancing,chen2023progressive}{, }\\
                        Replay Mechanisms~\cite{chen2023our,zhou2023replay}{, }\\
                        Meta-Knowledge Storage and Retrieval~\cite{sen2021rdfm,xu2023unleashing}{... }\\
                        ,fill=green!0
                        ]
                    ]
                    [
                        Model Interpretability and Transparency
                        [
                        Complex Internal Mechanisms~\cite{huff2021interpretation}{, }\\
                        Limited Data Availability~\cite{hoyos2023case}{, }\\
                        Development of Interpretable Architectures~\cite{hong2020human,hassija2024interpreting}{, }\\
                        Model Interpretation and Visualization Tools~\cite{montavon2019layer,achtibat2024attnlrp}{, }\\
                        Incorporating Domain Knowledge~\cite{smith2023biomedical}{, }\\
                        Indicators and Evaluation Techniques for\\ Interpretability~\cite{nauta2023anecdotal}{... }\\
                        ,fill=green!0
                        ]
                    ]
                    [
                        Model Bias and Fairness Issues
                        [
                        Identification of Bias Sources~\cite{drukker2023toward}{, }\\
                        Development of Fairness Metrics~\cite{jiang2023evaluating}{, }\\
                        Diverse and Representative Data Collection~\cite{singh2023unified}{, }\\
                        Bias Detection and Correction Techniques~\cite{prakhar2023bias}{, }\\
                        Fairness-aware Model Design~\cite{shi2023towards}{, }\\
                        Continuous Monitoring and Evaluation~\cite{wang2023automated}{... }\\
                        ,fill=green!0
                        ]
                    ]   
                ]
            ]
        ]
    \end{forest}
}
\end{adjustwidth}
\caption{An introduction to the image-text multimodal models, including its history, the methods and techniques currently available, as well as the challenges encountered and the direction of future development.}
\centering
\label{fig:1}
\end{figure*}

The image-text multimodal model has undergone the development process described above ~\cite{gan2022vision}. Currently, a unified paradigm has been formed to build high-quality datasets to enhance the fusion of graphic and textual information with a big language model as the core. There are several reviews available for image-text multimodal models, each with a different focus. Summaira \textit{et al}~\cite{Jabeen2022ARO} categorizes different modes based on their application and describes them in detail. Wang \textit{et al}~\cite{wang2023large} presents a comprehensive compilation of the latest algorithms used in multimodal large-scale modeling, along with the data sets used in recent experiments. Yin \textit{et al}~\cite{yin2023survey} in their review categorized and distinguished different types of multimodal algorithms in recent years. Wu \textit{et al}~\cite{wu2023multimodal} review the historical development of multimodal algorithms and discuss potential applications and challenges in this area. While previous reviews have examined image-text multimodal models in various contexts, there is limited literature that delves deeply into its applications in biomedical  domains.

The objective of this paper is to examine the evolution of generalized multimodal models and their influence on domain-specific applications in a novel manner. This paper will present a comprehensive analysis of the core concepts of image-text multimodal modeling and its historical development trajectory, with a particular focus on its innovative applications and breakthroughs in the biomedical field. This paper presents a summary of image-text multimodal models in terms of the dimensions of a task. It also demonstrates the excellent results achieved by image-text multimodal models in the biomedical field and cites authoritative literature to provide a solid academic foundation for these results. Furthermore, this paper will examine the shortcomings of the model and propose practical strategies for addressing these limitations in light of the latest research findings in the academic community. This will provide new insights and directions for research in related fields. Our research contributions are summarized in the following areas:

\begin{itemize}
\item We sort out the technological evolution of image-text multimodal modeling and illustrate how it contributes to the development of multimodal technologies in the biomedical field.
\item We analyzed the image-text multimodal model in depth in terms of the dimensions of the task and summarize the architecture, components and data of the generic model, as well as discuss the application and improvement of image-text multimodal models in the biomedical domain as an example.
\item We dissect the technical and applied challenges of mage-text multimodal models and provide targeted suggestions for future research through a structured analysis.
\end{itemize}

This paper will follow the structure presented in Figure \ref{fig:1}. In Section \ref{section2}, we will examine the fundamental concepts of image-text multimodal models, its developmental history,and its evolution in the biomedical field. Section \ref{section3} will demonstrate the performance of graphical macromodeling by presenting multiple tasks and explaining their unique manifestations in the biomedical domain. Section \ref{section4}, we will deeply probe into the challenges and limitations faced by image-text multimodal models, such as the requirement of extensive data and computational resources and the effective application of these models in resource-constrained environments. Following this, we will explore potential research directions for image-text multimodal models. This paper will conclude with a summary of the whole discussion in Section \ref{section5}.

\section{Development Process}
\label{section2}
\subsection{Technical Evolution of Image-text Multimodal Models}
\label{2.1}
\begin{figure*}[htbp]
    \centering
    \includegraphics[width=0.8\textwidth]{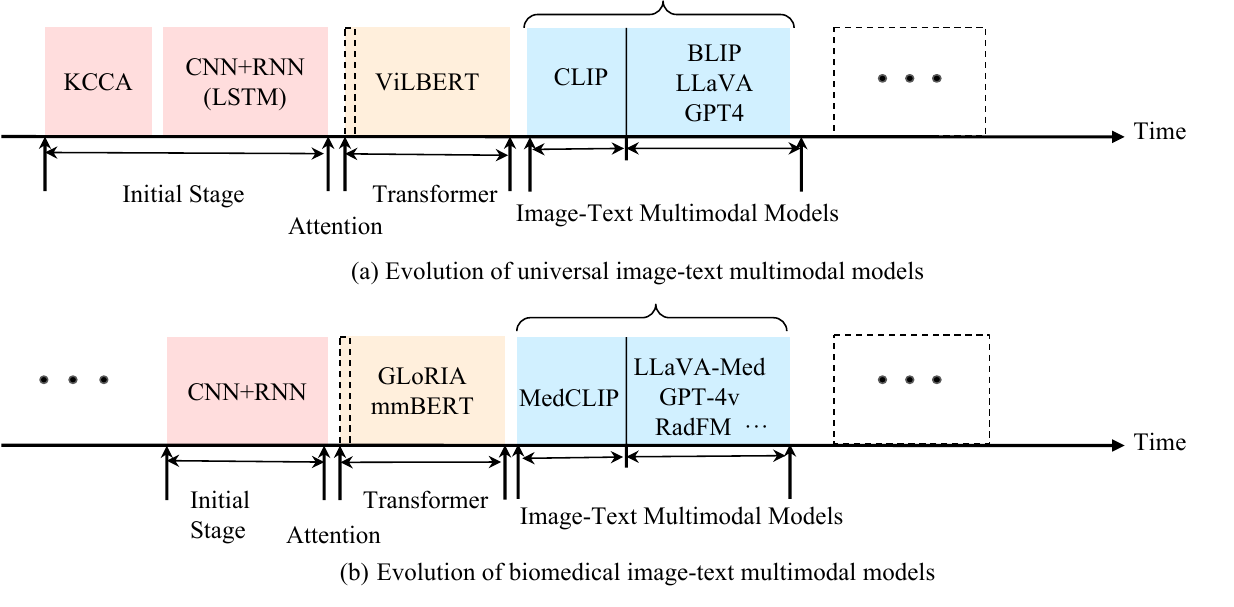}
    \caption{The Evalution of Some Image-Text Multimodal Models }
    \label{fig:2}
\end{figure*}

\begin{figure*}[htbp]
    \centering
    \includegraphics[width=0.8\textwidth]{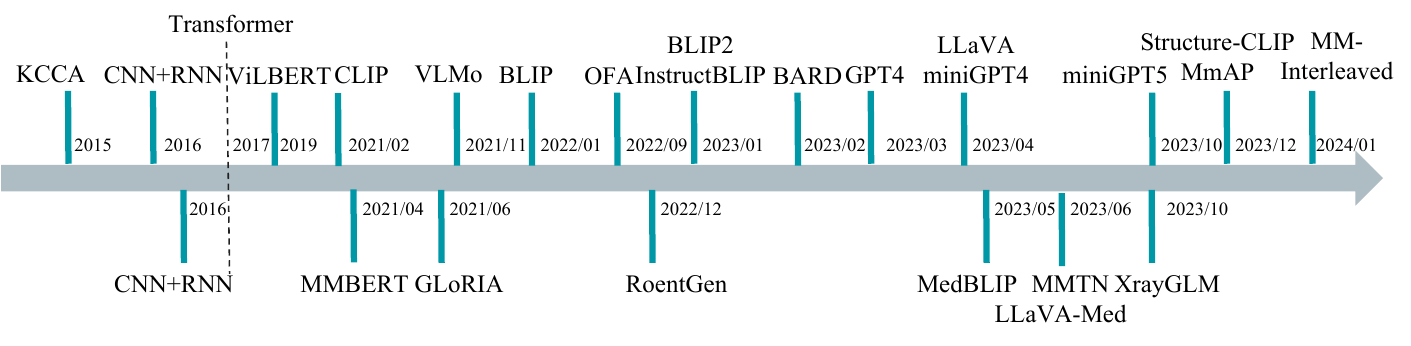}
    \caption{Evolution of representative image-text multimodal models}
    \label{fig:3}
\end{figure*}

The correspondence between image-text multimodal models and models applied in the biomedical field is shown in Figure \ref{fig:2}, as shown in (a) and (b) respectively. 
And the development process of the image-text multimodal models is shown in Figure \ref{fig:3}.

\subsubsection{Initial Stage and Early Stage}

Image-text multimodal models~\cite{baltruvsaitis2018multimodal,guo2019deep,gao2020survey,jiang2021review,suzuki2022survey,xu2023multimodal,cui2024survey,wang2024exploring}are a product of artificial intelligence  that can process and understand various types of information, such as text and images, simultaneously. These models aim to integrate multiple sensory inputs or data types into a unified framework to more comprehensively understand and process information. 
Image-text multimodal models are at the forefront of technological innovation, capable of processing diverse forms of input and, in certain instances, producing varied output modalities.
Training these multimodal models usually requires a large amount of annotated data, as they need to understand and process multiple types of data simultaneously. Collecting and annotating such extensive multimodal datasets can be both expensive and time-consuming.

In the initial stage, the primary research emphasis lies in  learning the joint feature space between images and their corresponding descriptions.
These methods project image features and sentence features into a common space, which can be used for image search or ranking image captions. Image-text multimodal models initially use various methods to learn projections, including N-grams methods ~\cite{Li2011ComposingSI}, Corpus-Guided methods ~\cite{Yang2011CorpusGuidedSG}, and kernel canonical correlation analysis (KCCA) methods~\cite{hodosh2013framing}.

The early stages of the development of image-text multimodal models focus on exploring the bidirectional mapping between images and text descriptions. Research initially attempts to use CNN and RNN. Because the CNN architecture benefits from the convolution kernel mechanism, it is particularly suitable for identifying and classifying local patterns in images. CNN is usually used to extract and process image features; RNN is usually used to process text sequence data due to its temporal processing mechanism. When generating or building a visual representation of the scene as the title is read dynamically, the representation automatically learns to remember long-term visual concepts, generate novel captions given images, and reconstruct visual features given image descriptions~\cite{chen2015mind,Mao2014DeepCW,mao2014explain,karpathy2015deep}.

Although the capabilities of early models are limited, they lay the foundation for subsequent multi-modal research. But RNN has weaknesses in re-memorizing concepts after several iterations. Without dedicated gating units, RNN language models often have difficulty learning long-distance relationships. Since RNN is not good at long-term memory, the output value range of the activation function is [0,1]. When information and residuals are transmitted in RNN neurons, they will be lost over time. LSTM has solved the problem of conventional RNN architecture's vanishing and exploding gradient problem.
Therefore, when applying feature conversion technology in the image field, it is necessary to convert the images in the video into fixed-dimensional vector representations and use LSTM recurrent neural networks to deeply model these vectors. In this process, LSTM has acted as a "decoder" that interprets these visual feature vectors representing video content and converts them into natural language descriptions. To solve the problem of requiring a large amount of supervised training data in this process, a strategy of transferring knowledge from auxiliary tasks is adopted. In this way, a single deep neural network can  convert video pixels into text output directly~\cite{venugopalan2014translating,venugopalan2015sequence,pan2016jointly,donahue2015long}.

\subsubsection{The Introduction of Attention Mechanism and the Rise of Transformers}

With the introduction of the attention mechanism, the capabilities of image and text multi-modal models improve significantly~\cite{yu2017multi,li2018multi,chen2017amc,baltruvsaitis2018multimodal,guo2019deep}. The attention mechanism allows the model to focus on specific parts of the image while generating relevant text descriptions ~\cite{lee2019attention,niu2021review,lu2023multi}. This method is more flexible and precise in processing images and text, and better captures the correlation between images and text.

On the basis of CNN and LSTM, the graphic description of the attention mechanism is introduced ~\cite{xu2015show} ~\cite{vinyals2015show}. A "soft" deterministic attention mechanism can be trained through the standard backpropagation method; a "hard" stochastic attention mechanism can be trained by maximizing approximate variational lower bounds, or equivalently, through reinforcement~\cite{zadeh2018memory}.

However, learning attention distribution for each pair of multi-modal input channels is expensive computationally. The Bilinear Attention Network (BAN) ~\cite{kim2018bilinear} finds bilinear attention distributions to seamlessly exploit given visual-linguistic information. BAN considers the bilinear interactions between the two sets of input channels, while low-rank bilinear pooling extracts a joint representation of each pair of channels.

The emergence of the Transformer model ~\cite{vaswani2017attention} marks an important milestone in the development of multi-modal models. Unlike RNN and CNN, Transformer relies on a self-attention mechanism to process sequence data, which makes it more efficient in processing long sequences. The Transformer model eliminates traditional loop and convolution operations and excels at handling long sequences and parallel computing. The introduction of Transformer improves the performance of image and text multi-modal models greatly, especially in language understanding.

The emergence of the BERT model ~\cite{devlin2018bert} has transformed the field of natural language processing. BERT uses two pre-training tasks: Masked Language Modeling (MLM) and Next Sentence Prediction (NSP) to effectively learn in-depth text representations. Notably, the influence of the Transformer model has extended beyond the NLP domain and has been applied to multimodal scenarios. Within those Transformer-based models, some mix image and text data right from the input stage, while others provide two separate encoders (or streams) for visual and textual information. Therefore, according to the model architecture, they can be divided into single-stream and dual-stream models. Common single-stream models include: VisualBERT ~\cite{li2019visualbert} integrates visual information into the text sequence in the form of specific tokens directly on top of BERT. Unicoder-VL ~\cite{li2020unicoder} is a universal vision-language pre-training model, which also employs a single transformer structure to handle visual and language inputs. VL-BERT ~\cite{DBLP:conf/iclr/SuZCLLWD20} is similar to VisualBERT, also processing visual and textual information in a single-stream manner. Oscar ~\cite{zhou2020unified} enhances the alignment of visual and textual information by using object tags as anchors. ALIGN ~\cite{jia2021scaling}, by pre-training on a large-scale dataset of images and noisy text, achieves outstanding performance on multiple downstream vision-language tasks. FILIP ~\cite{yao2021filip} aims to further strengthen the model’s ability to correlate image and text information by performing fine-grained contrastive learning. Common dual-stream models include: ViLBERT (Visual and Language BERT) ~\cite{lu2019vilbert}, which uses a dual-stream structure to separately process visual and textual information and implements their interaction through a co-attention mechanism at the top layer. LXMERT ~\cite{tan2019lxmert}, specifically designed for visual question answering tasks, uses a dual-stream architecture to encode images and text separately before interacting and merging them. ERNIE-ViL ~\cite{sun2019ernie} incorporates knowledge graphs into the semantic interaction of images and text, adopting a dual-stream structure. Pixel-BERT ~\cite{huang2020pixel}, unlike the original dual-stream design of ViLBERT, aligns the features of images and text directly at the pixel level, but the dual-stream feature processing is its characteristic. 

\subsubsection{Recent Image-text Multimodal Models}

Multimodal models has begun to adopt an end-to-end training method, which means that the model can learn the characteristics of images and text simultaneously within a single framework, rather than processing them separately. This approach enables the model to better capture the complex relationship between visual and textual information. The high degree of integration and collaboration of this model relies on two key components: text encoder and visual encoder.

The main responsibility of the text encoder module is to understand and process natural language text. By applying deep learning and natural language processing techniques, especially Transformer architecture-based models such as BERT and GPT, these encoders are able to deeply analyze the semantics and structure of text. They are not only able to identify keywords and phrases in text, but also understand more complex language structures and meanings, allowing the model to understand language information at a deeper level.

Meanwhile, the visual encoder module focuses on extracting and understanding visual information from images or videos ~\cite{li2019visual,khan2022transformers,liu2023survey}. These encoders are usually based on convolutional neural networks (CNN) or Transformer-based vision models that have emerged in recent years, such as Vision Transformer (ViT). Their function is to capture elements such as objects, textures, colors, etc. in the image and convert this visual information into a form that allows the model to be further processed and analyzed.

Image-text multimodal models combine information from image and text modalities. For example, CLIP ~\cite{radford2021learning}, a visual encoder models image information, while a text encoder models text information. The feature vectors of images and text can be pre-calculated and stored, and the modal interaction is handled through the cosine similarity of the image and text feature vectors. Since CLIP was proposed, it has become a classic in the field of multimodal learning and is widely used as the basic model of today's multimodal models. CLIP uses 400 million pairs of (image, text) data for training through self-supervision. It maps images and text into a shared vector space, allowing the model to understand the semantic relationship between images and text. This approach represents a new way to learn visual models under natural language supervision. This type of model can better handle multi-modal retrieval tasks but cannot handle complex classification tasks. However, ViLT ~\cite{kim2021vilt} has only a visual-text encoder that models image information and text information simultaneously, using the Transformer model to interact with image and text features. ViLT is inspired by the patch projection technology in ViT ~\cite{dosovitskiy2020image}. It aims to minimize the feature extraction of each modality, using pre-trained ViT to initialize the interactive Transformer, and directly uses the interactive layer to process visual features without adding new layers. ViLT's visual encoder focuses the main computational effort on the feature fusion part of the transformer. This type of model can fully integrate multi-modal information, is better at multi-modal classification tasks, and is slower at retrieval tasks.
CLIP focuses on the matching and association between images and text, while ViTL focuses more on image recognition and classification tasks. In view of the advantages and disadvantages of CLIP and ViTL, VLMo ~\cite{bao2022vlmo} proposed a unified visual language model-Mixture-of-Modality-Experts (MoME). After the VLMo pre-training is completed, it can be used in a double-tower structure to efficiently retrieve images and texts; or in a single-tower structure to interact with the features of fused images and texts.

With the exponential growth of large-Scale language models and the continuous emergence of new models, the field of artificial intelligence is undergoing an unprecedented revolution. These models, characterized by their massive parameter scales and advanced architectures, exhibit remarkable performance in natural language processing and text generation tasks. They excel in a wide array of applications, from automated question answering to machine translation, and even conversational systems. BLIP ~\cite{Li_Li_Xiong_Hoi} can flexibly switch between visual-language understanding tasks and generation tasks, and effectively utilize noisy data through bootstrapping. BLIP-2 ~\cite{Li_Li_Savarese_Hoi} introduces a lightweight Querying Transformer (Q-former) that connects visual large-scale models and language large-scale models. Q-former is trained in two stages: fixing the image encoder to learn representations of visual-language consistency, and fixing the language large-scale model to enhance the generation capability from vision to language. InstructBLIP~\cite{dai2305instructblip} proposes a paradigm of instruction fine-tuning based on BLIP-2, using additional instructions to extract more useful visual features. mPLUG-2~\cite{xu2023mplug} is capable of handling various modal inputs such as text, images, and videos. It has achieved leading or comparable performance in more than 30 multimodal and unimodal tasks. Models such as BEIT3~\cite{wang2022image} and X-FM~\cite{chen2022learning} undergo joint pre-training on both image and text data, enabling them to learn joint representations of images and text. These models can handle not only tasks that involve only text or images but also complex tasks that require simultaneous processing of both text and images, such as generating image descriptions and answering questions related to images. LLaVA ~\cite{Liu_Li_Wu_Lee_-Madison_Research} fine-tunes on the generated visual-language instruction data end-to-end, validating the effectiveness of using generated data for instruction-tuning in LMM, and providing practical techniques for constructing general instructions that follow visual agents. OFA~\cite{wang2022ofa} unifies the understanding and generation tasks of multimodal and single-modal data into a single simple Seq2Seq generative framework. It undergoes pretraining and fine-tuning using task instructions, without introducing additional task-specific layers. GPT-4 ~\cite{bubeck2023sparks}takes this to the next level, boasting a total parameter count of 1.8 trillion, a significant increase compared to GPT-3's 100 billion parameters. This massive parameter scale equips GPT-4 with potent computational capabilities and potential in natural language understanding and generation tasks. It enhances its contextual comprehension, allowing it to handle more complex textual information and possess advanced semantic analysis and knowledge reasoning abilities. MiniGPT-4 ~\cite{zhu2023minigpt} fixes the parameters of the language model and the visual model, and then uses only the projection layer to align the language decoder with LLaMA and the visual encoder with BLIP-2. LRV-Instruction effectively reduces hallucinations produced by LMMs, and maintaining a balanced ratio of positive and negative instances leads to a more robust model~\cite{liu2023aligning}. MiniGPT-5 ~\cite{zheng2023minigpt} combines the Stable Diffusion mechanism with LLM through special visual tokens. In addition to its original multi-modal understanding and text generation capabilities, it can also provide reasonable and coherent multi-modal output. Structure-CLIP ~\cite{huang2023structure}proposes a Knowledge Enhanced Encoder (KEE) that takes SGK as input to enhance multi-modal structured representation by integrating scene graph knowledge (SGK). In order to maximize the complementarity of highly similar tasks, CLIP-based MmAP~\cite{xin2023mmap} utilizes a gradient-driven task grouping method to divide tasks into several disjoint groups and assign a group-shared parameter to each group. MM-Interleaved ~\cite{tian2024mm} achieves direct access to fine-grained image features in the previous context during the generation process by introducing a multi-scale multi-image feature synchronizer module. This module enables models to be pre-trained end-to-end on paired and interleaved image-text corpora to more efficiently follow complex cross-modal instructions.

In the real world, data are not always complete, and situations where certain modalities are missing may be encountered. This results in degraded model performance. Integrating the missing-aware prompt method into Transformer-like multimodal models is straightforward, which can alleviate the impact of missing modalities and enhance the robustness of the models~\cite{lee2023multimodal, ma2022multimodal, ma2021smil, zeng2022tag, zhao2021missing}.

The technical development of large multimodal models for vision and text has gone through many important stages, and it has essentially formed a unified paradigm that uses large language models as the core and leverages high-quality datasets to enhance the fusion of visual and textual modalities. With the development of multimodal technology, modal fusion occupies a core position in the technological advancement of large multimodal models for vision and text. This is because information from different modalities is complementary, and their effective combination can improve the models' ability to understand complex phenomena. Multimodal models have been able to more understand deeply and process the relationship between images and text, which is not limited to surface-level combinations but involves deep semantic and contextual understanding.

\subsection{Evolution of Application Technology in the Image-text Multimodal Models on the biomedicine}

In the general field, image-text data are very abundant, and multimodal pre-trained models in the field of natural images and text, such as multi-modal GPT-4 and LLaVA, have demonstrated remarkable results on various tasks. However, applying these models to the biomedical domain directly is fraught with challenges due to the particularity and complexity of biomedical data, which complicates cross-domain migration. Additionally, obtaining biomedical image datasets requires significant investment and expertise.

In the early stages, Convolutional Neural Networks (CNNs) were first utilized independently in the biomedical domain ~\cite{salehi2023study,sarvamangala2022convolutional,yao2022comprehensive}, due to their powerful feature processing capability, local receptive fields, and the characteristic of parameter sharing, which eliminated the need for manual design or selection of features. Primarily, they were applied to the segmentation of biomedical images. Gheisari et al. adopted a combined approach of CNNs and Recurrent Neural Networks (RNNs) ~\cite{gheisari2021combined} to better extract spatial and temporal information, improving the accuracy of glaucoma prediction significantly. However, CNNs may face many challenges when dealing with biomedical images that have complex structures and correlations, due to the complexity of hyperparameter selection and limitations in learning long-range dependencies.

CNNs have limited capability in handling long-range dependencies, as they extract features through local receptive fields and struggle to capture long-distance patterns in the data. It is these limitations that have prompted the birth and development of the attention mechanism. The attention mechanism can give the network "focus points" when processing input data, allowing the model to concentrate on the most relevant parts of the image while ignoring other less important information dynamically. This mechanism mimics the focusing ability of the human visual system and is particularly valuable in biomedical image analysis, as it can aid models in more identifying pathological areas accurately or carrying out fine-grained structural segmentation.

The introduction of the attention mechanism has also led to new technological innovations in biomedicine ~\cite{zhou2021review,aminizadeh2023applications}. The Transformer model, as a pioneering application of the attention mechanism, has been  used widely in the field of natural language processing and has demonstrated significant advantages. Entirely based on the attention mechanism, the Transformer does not rely on recurrent networks to capture long-distance dependencies in input sequences, enabling parallel computation and greatly enhancing training efficiency. Additionally, the attention mechanism offers improved interpretability and allows the Transformer to more flexibly focus on key information in the input data. However, despite the innovations and advantages brought by the Transformer, it has a large number of parameters and requires more computational resources for effective training. GLoRIA ~\cite{huang2021gloria} utilizes an attention mechanism to learn global-local representations of images by matching words and image sub-regions in radiology reports, creating context-aware local image representations by learning attention weights based on important image sub-regions of specific words. MMBERT ~\cite{khare2021mmbert} achieves new state-of-the-art performance on two biomedical Visual Question Answering (VQA) datasets by leveraging existing large multimodal biomedical datasets to learn better image and text representations.

LViT ~\cite{li2023lvit} is used for biomedical image segmentation mainly. It is a double U structure, consisting of a U-shaped CNN branch and a U-shaped Transformer branch. The CNN branch is responsible for image input and prediction output, the ViT branch is used to merge image and text information, and uses Transformer to process cross-modal information. REFERS~\cite{zhou2022generalized} uses transformer to extract relevant feature representations of different views. To utilize the information of each report fully, an attention mechanism-based view fusion module is designed to process all radiographs in a patient study and fuse multiple features simultaneously. O-Net~\cite{wang2022net} integrates CNN and Transformer ingeniously, two widely favored technologies in the field of image processing, aiming to enhance performance in biomedical image segmentation and classification. It is capable of leveraging both global and local characteristics of images to provide a more accurate and detailed analysis fully. In the encoder part, O-Net utilizes both CNN and Swin Transformer to identify and focus on global and local contextual features within the images; whereas in the decoder part, O-Net merges the outcomes from Swin Transformer and CNN to achieve the final image segmentation or classification. This architecture is particularly beneficial in the general visual domain, which often requires models to understand interactions between different objects. The Transformer structure is advantageous for modeling long-distance dependencies, enhancing the model's ability to understand the entire scene, which is useful for complex image comprehension especially.
FTransCNN ~\cite{ding2023ftranscnn} consists of CNN and Transformer architectures and employs a novel fuzzy fusion strategy to utilize features extracted by CNNs and Transformers jointly. As parallel feature extraction backbones, channel attention is used to enhance the global key information in the Transformer for improved feature representation, while spatial attention boosts the local details in CNN features and suppresses irrelevant areas.  The use of multiple models for extracting segmentation features also avoids heterogeneity and uncertainty. DesTrans ~\cite{song2024destrans} integrates Transformer and DenseNet network modules to address issues of edge blurring and information redundancy in biomedical image fusion, thereby enhancing the accuracy of disease diagnosis. When applied to natural images, the Transformer and densely cascaded model structure use restricted feature loss to enhance the method's feature extraction capabilities, thus reducing the risk of edge blurring.

However, models developed in the biomedical field perform poorly when applied to general domains directly. This is because biomedical datasets face several unique challenges that hinder the performance of models trained on biomedical applications in general image-text domains ~\cite{qiu2023pre}. 
\textbf{Data Scarcity and Annotation Difficulty.} The acquisition and annotation of biomedical imaging data are extremely costly, requiring the involvement of professionals such as radiologists. This not only increases the difficulty of obtaining datasets but also results in relatively small dataset sizes. In general image-text domains, data is more abundant typically and easier to annotate. For example, the ImageNet dataset contains millions of well-annotated images, whereas the largest datasets in the biomedical field usually have only hundreds of thousands of images. 
\textbf{High Specialization of Biomedical Data.} Biomedical images contain highly specialized information that requires expert knowledge to interpret and analyze. Models trained on such data have highly specialized feature extraction capabilities, but these features may not be applicable in general image-text domains. For instance, a model trained to identify lung nodules would be of no practical use for ordinary photographs, as the features extracted by these models are unrelated to general object recognition.
\textbf{Privacy and Data Sharing Restrictions.} Biomedical data involves patient privacy and is strictly regulated by legal and ethical standards. This limits the public sharing and broad use of such data, resulting in a lack of diversity in training data and subsequently affecting the model's generalization ability. In general image-text domains, data sharing is more open and widespread, allowing models to learn from a more diverse set of data.

With the large-scale image and text pre-training model, CLIP has achieved considerable success in the field of biomedicine. MedCLIP ~\cite{wang2022medclip} decouples picture-text pairs effectively and performs comparative learning, reducing false negatives by introducing external biomedical knowledge significantly. Chambon et al.~\cite{chambon2022adapting} have utilized generative models to generate biomedical images, helping to alleviate the shortage of biomedical datasets. Their main research focuses on extending the representation capabilities of large-scale pre-trained base models to encompass biomedical concepts, enabling the generation of images that match the semantics of text closely and distinguish the presence or absence of conditions such as "pleural effusion". RoentGen~\cite{chambon2022roentgen}, a generative model, synthesizes high-fidelity chest radiographs. It allows for the insertion, combination, and modification of various chest radiograph images through free-form biomedical language text prompts, demonstrating high correlation with corresponding biomedical concepts. MMTN~\cite{cao2023mmtn} integrates and processes multimodal biomedical data, including biomedical images, terminology knowledge, and text reports. It explores interactions between different modalities to enhance the quality of biomedical report generation, achieving reports that are consistent with real-world observations.

General-domain graphic-text multimodal assistants might not accurately respond to biomedical inquiries, providing incorrect or misleading information potentially. LLaVA-Med ~\cite{li2023llava} represents the inaugural effort to adapt multimodal instruction techniques specifically for the biomedical domain, employing end-to-end training to develop biomedical multimodal dialogue assistants. XrayGLM ~\cite{wang2023XrayGLM} demonstrates significant potential in biomedical imaging diagnosis and multi-round interactive dialogue.

Transitioning to the evaluation of multimodal biomedical diagnosis, Chaoyi Wu et al. utilized GPT-4 to assess 17 human body systems and observed its proficiency in identifying biomedical image modalities and anatomy. However, it encountered difficulties in disease diagnosis and generating comprehensive reports, highlighting major challenges in the field~\cite{wu2023can}. RadFM~\cite{wu2023towards} has achieved state-of-the-art (SOTA) performance in five tasks, including modal recognition, disease diagnosis, visual question answering, report generation, and principal diagnosis. This underscores the model's comprehensive capability to address real clinical challenges.

In conclusion, the application technology of image-text multimodal models in biomedicine has mirrored the broader technical evolution of multimodal models, as outlined in Section \ref{2.1}. The journey from initial explorations to the current state-of-the-art applications has been marked by significant milestones, each contributing to a more integrated and sophisticated understanding of multimodal data. In biomedicine, this evolution has been impactful particularly, as it leverages the core advancements in large language models and high-quality datasets to address the unique challenges and complexities of biomedical data. The fusion of visual and textual modalities in this domain is not merely a technical achievement but a transformative approach to enhancing diagnostic accuracy, patient care, and biomedical research.  
As illustrated in Figure 2, the evolution trajectory highlights a trend towards models that not only merge but also integrate effectively and interpret multimodal information. This trend indicates the potential for incremental advancements in biomedicine. The interplay between the technical developments in general multimodal models and their specific applications in biomedicine demonstrates a complementary progression. 


\section{Applications of image-text multimodal models In image-text tasks}
\label{section3}
We will introduce multimodal models based on image-text mode. Image-text data typically possesses clearer semantic information because text provides precise descriptions while images offer intuitive visual information. Additionally, image-text data usually has a smaller volume, making processing and analysis more efficient, thereby leading to a plethora of research~\cite{khan2022single,yang2020image,huang2019image} focusing on the image-text modality. Furthermore, the fusion of image and text information in image-text data provides a more comprehensive and multi-perspective view of the data, aiding in accurate and holistic data analysis.
In this chapter, we first analyze three specific field tasks of image-text multimodal models and analyze their common components and challenges. Then, we summarize the architecture, components and data of the general image-text multimodal model. Finally, we take biomedicine as an example to introduce the application and improvement of image-text multimodal models in other fields.
\subsection{Tasks}

The image-text multimodal models mainly involve image caption, image-text matching, and Visual Question Answering (VQA) tasks. They represent different types of relationships and interactions between images and text. The image caption task requires the model to generate natural language text describing the content of an image, while the image-text matching task requires the model to understand the semantic relationship between images and text. Additionally, the VQA task further examines the model's comprehensive understanding of both images and text by posing questions related to images and evaluating the model's performance based on its ability to answer these questions. Therefore, these three tasks cover various interaction modes between images and text, making them representative for evaluating the comprehensive capabilities of image-text multimodal models.
\subsubsection{Pre-training Task}
The confluence of computer vision and natural language understanding has emerged as a prominent research focus. We focus on three tasks: Image Caption Generation, Image-text Matching, Visual Question Answering (VQA). These tasks facilitate enhanced comprehension of natural language, extraction of desired information from imagery. The proliferation of image-text multimodal models has catalyzed advancements within these domains. 
The evaluation indexes corresponding to the three tasks of the multimodal model are shown in Table \ref{metric}, and the evaluation indexes corresponding to the classical model are shown in Table \ref{metricvalue}.

\begin{table}[t]
\footnotesize
    \centering
    \caption{Common metrics in three tasks of multimodal model.}
    \renewcommand{\arraystretch}{1.2}
    \begin{tabular}{>{\centering\arraybackslash}p{3cm}>{\centering\arraybackslash}p{3cm}p{3cm}}
        \toprule
        Task & Metric \\
        \midrule
        \multirow{5}{*}{Image Caption} & BLEU-n \\
        & CIDER \\
        & METEOR \\
        & ROUGE-L \\
        & SPICE \\
        \midrule
        \multirow{3}{*}{Image-text matching} & Recall@1 \\
        & Recall@5 \\
        & Recall@10 \\
        \midrule
        \multirow{5}{*}{Visual Question Answering} & Accuracy \\
        & CIDER \\
        & BLEU-n \\
        & ROUGE-L \\
        & Hit@K \\
        \bottomrule
    \end{tabular}
\label{metric}%
\end{table}

\begin{table*}[h]
\footnotesize
    \centering
    \caption{The metric value of the classical model in the three tasks of the multimodal model.}
    {\renewcommand\baselinestretch{1.25}\selectfont
\resizebox{1.0\textwidth}{!}{
    \begin{tabular}{ccccccc}
        \toprule
        Task & DataSet & Model & BLEU-4 & CIDER & METEOR & SPICE \\
        \midrule
        \multirow{8}{*}{\centering Image Caption} & \multirow{8}{*}{\centering COCO Captions~\cite{chen2015microsoft}} & BLIP-2 VIT-G OPT 2.7B (zero-shot)~\cite{li2023blip} & 43.7 & 145.8 & - & - \\
        & & BLIP-2 VIT-G OPT 6.7B (zero-shot)~\cite{li2023blip} & 43.5 & 145.2 & - & - \\
        & & BLIP-2 VIT-G FlanT5 XL (zero-shot)~\cite{li2023blip} & 42.4 & 144.5 & - & - \\
        & & mPLUG~\cite{li2022mplug} & 46.5 & 155.1 & 32 & 26 \\
        & & OFA~\cite{wang2022ofa} & 44.9 & 154.9 & 32.5 & 26.6 \\
        & & GIT~\cite{wang2022git} & 44.1 & 151.1 & 32.2 & 26.3 \\
        & & LEMON~\cite{hu2022scaling} & 42.6 & 145.5 & 31.4 & 25.5 \\
        & & ExpansionNetv2~\cite{hu2023exploiting} & 42.7 & 143.7 & 30.6 & 24.7 \\
        \midrule
        Task & DataSet & Model & Recall@1 & Recall@5 & Recall@10 & -\\
        \midrule
        \multirow{9}{*}{\centering Image-to-Text Retrieval } & \multirow{9}{*}{\centering Flickr30k~\cite{chen2015microsoft} } & BLIP-2 VIT-L (zero-shot, 1K test set)~\cite{li2023blip} & 96.9 & 100 & 100 & - \\
        & & BLIP-2 VIT-G (zero-shot, 1K test set)~\cite{li2023blip} & 97.6 & 100 & 100 & -\\
        & & InternVL-G-FT~\cite{chen2023internvl} & 97.9 & 100 & 100 & -\\
        & & ONE-PEACE~\cite{wang2023one} & 97.6 & 100 & 100 & - \\
        & & InternVL-C-FT & 97.2 & 100 & 100 & - \\
        & & ERNIE-ViL 2.0~\cite{shan2022ernie} & 96.1 & 99.9 & 100 & - \\
        & & ALBEF~\cite{li2021align} & 95.9 & 99.8 & 100 & - \\
        & & GSMN~\cite{nguyen2021deep} & 76.4 & 94.3 & 97.3 & - \\
        & & LSGCM~\cite{nguyen2021deep} & 71 & 91.9 & 96.1 & - \\
        \midrule
        Task & DataSet & Model & Acuracy & - & - & -\\
        \midrule
        \multirow{12}{*}{\centering VQA} & \multirow{12}{*}{\centering GQA test-dev~\cite{hudson2019gqa}} & BLIP-2 VIT-G FlanT5 XL (zero-shot)~\cite{li2023blip} & 44.2 & - & - & -\\
        & & BLIP-2 VIT-G FlanT5 XXL (zero-shot)~\cite{li2023blip} & 44.7 & - & - & -\\
        & & BLIP-2 VIT-L OPT 2.7B (zero-shot)~\cite{li2023blip} & 33.9 & - & - & - \\
        & & BLIP-2 VIT-G OPT 2.7B (zero-shot)~\cite{li2023blip} & 34.6 & - & - & - \\
        & & BLIP-2 VIT-G OPT 6.7B (zero-shot)~\cite{li2023blip} & 36.4 & - & - & - \\
        & & BLIP-2 VIT-L FlanT5 XL (zero-shot)~\cite{li2023blip} & 36.4 & - & - & - \\
        & & CFR~\cite{nguyen2022coarse} & 72.1 & - & - & - \\
        & & PaLI-X-VPD~\cite{hu2023visual} & 67.3 & - & - & -\\
        & & NSM~\cite{hudson2019learning} & 63 & - & - & -\\
        & & Lyrics~\cite{lu2023lyrics} & 62.4 & - & - & -\\
        & & LXMERT~\cite{tan2019lxmert} & 60 & - & - & -\\
        & & single-hop + LCGN~\cite{hu2019language} & 55.8 & - & - & - \\
        \bottomrule
    \end{tabular}}}
\label{metricvalue}%
\end{table*}

\textbf{Image caption generation}
\begin{figure*}[ht]
    \centering
    \includegraphics[width=0.8\textwidth]{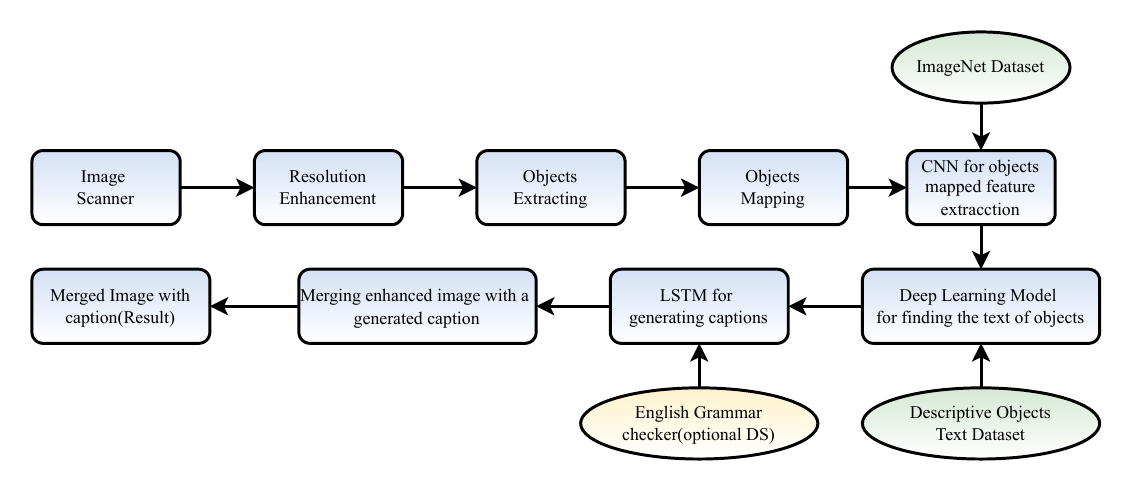}
        \caption{Image Captioning Process in Modern Techniques.}
    \label{fig:3-1}
\end{figure*}

The generation of image descriptions was initially proposed by Farhadi, Hejrati, et al.~\cite{farhadi2010every} using a template matching approach as shown in Figure \ref{fig:3-1}. This method involves solving Markov Random Field (MRF) images to predict triplets (object, action, scene). Subsequently, the template is filled with the given triplets to generate a descriptive image. Manual addition of descriptions and labels to the dataset is required to create own dataset. Later, retrieval-based methods were developed, involving the analysis of input images and comparing the similarity between input images and a known image database. However, retrieval-based methods usually collect numerous samples that are often necessary to maintain the massive image database~\cite{al2022image,bai2018survey,stefanini2022show,kang2022survey,thakareautomatic}.

With the emergence of deep learning neural networks and big data, neural network-based approaches have been applied to image captioning tasks, such as encoding and decoding models. Some researchers use CNN models as image encoders and RNN models as decoders to generate sentences ~\cite{amritkar2018image,suresh2022image,ghandi2023deep}. However, RNN models suffer from the limitation of vanishing gradients, remembering only a limited time slice of content. To achieve good performance, feature extraction is based on CNN models ~\cite{gu2017empirical,kinghorn2018region}, later transitioning to using LSTM models to incorporate global feature information for generating corresponding text descriptions ~\cite{li2017image,tan2019phrase,liu2020image}. After introducing attention mechanisms, the feature representation of important regions in the current moment's image serves as a context vector input to the encoding-decoding model ~\cite{anderson2018bottom,xiao2019dense,huang2019attention,niu2021review,luo2022thorough}. Performance in image captioning tasks significantly improves by inputting features containing positional information to the model.

 The advent of large pre-trained models has led to significant breakthroughs in image caption generation ~\cite{hu2022scaling,xia2021xgpt,shah2023lm,wang2023large}. For example, Transformer-based image-text multimodal models, such as CLIP, BLIP, and BLIP-2, unify the encoding of image and text information, achieving excellent performance in various image-text tasks through pre-training and fine-tuning processes. Vision-Language Pre-training (VLP) methods enrich and enhance the accuracy of image descriptions generated by learning visual and language knowledge through pre-training ~\cite{wu2023medklip,wang2023learning,yin2023givl,ghandi2023deep}.

\begin{figure*}[ht]
    \centering
    \includegraphics[width=0.8\textwidth]{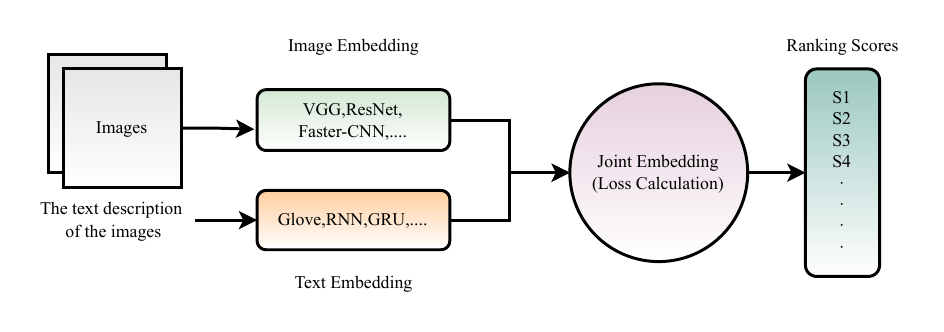}
    \caption{The basic structure for image-text matching task.}
    \label{fig:3-2}
\end{figure*}

\textbf{Image-text Matching}
Image-text matching refers to the task of jointly modeling images and text to identify semantic associations or similarities between them as shown in Figure \ref{fig:3-2}. The primary goal of this task is to ascertain whether images and text are semantically matched or related~\cite{diao2021similarity,chen2022two,zhang2022show}. Pre-trained large-scale models have significantly advanced the field of image-text matching, leading to notable progress. The methodologies in this domain can be broadly categorized into three types: contrastive learning-based image matching~\cite{yang2021tap,liang2021contrastive,xing2022neural}, attention mechanism-based generative model for image matching~\cite{lee2018stacked,wei2022bsam,diao2021similarity,maleki2022lile}, and generative model-based image matching~\cite{zhu2022image,li2022fine,zhao2023generative,tan2021cross,cheng2022vision}.

Utilizing a contrastive learning approach, researchers have developed frameworks that encode images and text into low-dimensional vectors. These models are trained to maximize the similarity of matched pairs while minimizing the similarity of unmatched pairs~\cite{wang2022coder,habib2023gac,ji2023knowledge}. This strategy enables precise matching of images and text, thereby enhancing the efficacy of image retrieval and related applications.

The attention mechanism-based image matching method dynamically adjusts attention weights based on the correlation between images and text, facilitating a deeper understanding of their semantic relationship~\cite{lee2018stacked,wang2019position}. This approach not only heightens the accuracy of image-text matching but also boosts the performance of tasks such as image annotation and image-centric question answering. Contemporary studies~\cite{alayrac2022flamingo} frequently employ cross-attention to emphasize relevant elements within images or text, aiming to synergize them. Some studies~\cite{fu2021cross} suggest incorporating self-attention as an additional loss term to enrich the representations within the cross-attention module.

Furthermore, generative model-based image-text matching involves creating textual descriptions for given images~\cite{fedus2018maskgan,tewel2021zero}. These models imbue images with richer semantic information, facilitating more precise matching with textual descriptions.

\begin{figure}[ht]
    \centering
    \includegraphics[width=0.45\textwidth]{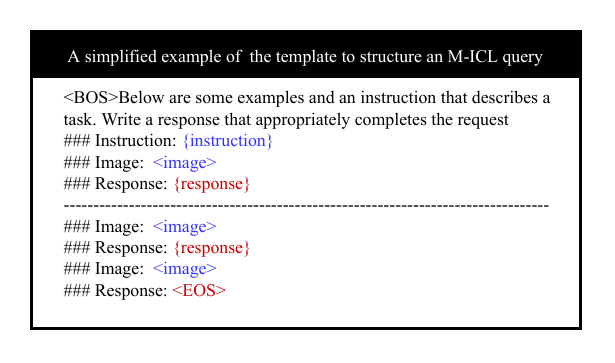}
    \caption{The basic structure for VQA task(adapted~\cite{gong2023multimodal}).}
    \label{fig:3-1-3}
\end{figure}

\textbf{VQA/Visual reasoning}
Visual Question Answering (VQA) and Visual Reasoning are a task aimed at enabling computers to answer natural language questions about an image. VQA usually needs to extract the image features and the query features and then integrate them to generate the answer as shown in Figure \ref{fig:3-1-3}
~\cite{cheng2022visual,sampat2022reasoning,malkinski2022deep}.

In recent years, with the development of deep learning and pre-training models, the performance of VQA and Visual Reasoning tasks has significantly improved. Large pre-training models have achieved remarkable results in VQA and Visual Reasoning tasks. These models leverage a massive amount of annotated data, enabling them to understand the intricate relationships between images and text. For example, Kant et al.~\cite{kant2021contrast} use contrastive learning approaches and are trained on a large scale of image-text pairs, effectively enhancing the model's performance in VQA tasks.

Researchers are also exploring more efficient model architectures and training strategies~\cite{chen2020hcp,kant2021contrast,lu2021transformer,qian2022pointnext}. For instance, certain methods employ attention mechanisms to capture the mutual relations between images and questions, thereby generating more accurate answers. Other approaches focus on how to carry out a reasoning process, such as through Graph Neural Networks ~\cite{yusuf2022evaluation,yusuf2022analysis,senior2023graph} or Transformer-based models ~\cite{khare2021mmbert,biten2022latr,seenivasan2022surgical}, to solve more complex visual reasoning tasks. In addition, Gao et al.~\cite{gao2021structured} point out that current models suffer a lack of textual-visual reasoning capability. To address this issue, the authors propose a novel end-to-end structured image-text multimodal attention neural network, which encodes the relationships between objects and text in images using a structural graph representation, and design a image-text multimodal graph attention network to reason over queries. This method achieved an accuracy of 45\% on the TextVQA~\cite{singh2019towards} dataset. mPLUG~\cite{li2022mplug} achieves effective and efficient visual-linguistic learning through novel cross-modal skip connections to address fundamental information asymmetry issues. Instead of fusing visual and language representations at the same level, mPLUG facilitates fusion between different abstraction levels at asymmetric levels through cross-modal skip connections.

\begin{table}[htbp]
\footnotesize
  \centering
  \caption{Comparison of general model architectures and biomedical models.}
    \begin{tabular}{ccr}
    \toprule
    \multicolumn{1}{c}{Task} & \multicolumn{1}{c}{Components} & \multicolumn{1}{l}{Representative Work} \\
    \midrule
    \multirow{23}[2]{*}{General Model} & \multicolumn{1}{c}{\multirow{12}[2]{*}{Architecture}} & \multicolumn{1}{l}{
          CLIP,~\cite{radford2021learning}}\\
           &       & \multicolumn{1}{l}
          {ViLT,~\cite{kim2021vilt}} \\
          &       & \multicolumn{1}{l}
          {BLIP,~\cite{Li_Li_Xiong_Hoi}} \\
          &       & \multicolumn{1}{l}
          {BLIP2,~\cite{li2023blip}} \\
          &       & \multicolumn{1}{l}
          {LLAVA-PLUS,~\cite{liu2023llava}} \\
          &       & \multicolumn{1}{l}
          {MiniGPT4,~\cite{zhu2023minigpt}} \\
          &       & \multicolumn{1}{l}
          {PandaGPT,~\cite{su2023pandagpt}} \\
          &       & \multicolumn{1}{l}
          {MobileVLM,~\cite{chu2023mobilevlm}} \\
          &       & \multicolumn{1}{l}
          {MobileVLM V2,~\cite{chu2024mobilevlm}} \\
          &       & \multicolumn{1}{l}
          {SpatialVLM,~\cite{chen2024spatialvlm}} \\
          &       & \multicolumn{1}{l}
          {MiniGPT-v2,~\cite{chen2023minigpt}} \\
          &       & \multicolumn{1}{l}
          {Qwen-VL,~\cite{bai2023qwen}} \\
          \cmidrule{2-3}
& \multirow{13}[0]{*}{Data} & \multicolumn{1}{l}{Open Images V7,~\cite{benenson2022colouring}} \\
          &       & \multicolumn{1}{l}{GQA,~\cite{hudson2019gqa}} \\
          &       & \multicolumn{1}{l}{Open Images V4,~\cite{kuznetsova2020open}} \\
          &       & \multicolumn{1}{l}{Fashion IQ,~\cite{wu2021fashion}} \\
          &       & \multicolumn{1}{l}{Conceptual 12M,~\cite{changpinyo2021conceptual}} \\
          &       & \multicolumn{1}{l}{SBU Captions,~\cite{ordonez2011im2text}} \\
          &       & \multicolumn{1}{l}{FLAVA,~\cite{singh2022flava}} \\
          &       & \multicolumn{1}{l}{UNIMO,~\cite{li2020unimo}} \\
          &       & \multicolumn{1}{l}{UNIMO-2,~\cite{li2022unimo}} \\
          &       & \multicolumn{1}{l}{LAION-400M/5B,~\cite{schuhmann2022laion}} \\
          &       & \multicolumn{1}{l}{WenLan,~\cite{huo2021wenlan}} \\
          &       & \multicolumn{1}{l}{RedCaps,~\cite{desai2021redcaps}} \\
          &       & \multicolumn{1}{l}{WIT,~\cite{srinivasan2021wit}} \\
    \midrule
    \multirow{25}[2]{*}{Medical Model} & \multirow{12}[1]{*}{Architecture} & \multicolumn{1}{l}{PMC-CLIP,~\cite{lin2023pmc}} \\
          &       & \multicolumn{1}{l}{MedBLIP,~\cite{chen2023medblip}} \\
          &       & \multicolumn{1}{l}{LLaVA-Med,~\cite{li2023llava}} \\
          &       & \multicolumn{1}{l}{Med-Flamingo,~\cite{moor2023med}} \\
          &       & \multicolumn{1}{l}{PMC-VQA,~\cite{zhang2023pmc}} \\
          &       & \multicolumn{1}{l}{MUMC,~\cite{li2023masked}} \\
          &       & \multicolumn{1}{l}{PMC-CLIP,~\cite{lin2023pmc}} \\
          &       & \multicolumn{1}{l}{PMC-LLaMA,~\cite{wu2023pmc}} \\
          &       & \multicolumn{1}{l}{MedAlpaca,~\cite{li2023llava}} \\
          &       & \multicolumn{1}{l}{Meditron,~\cite{chen2023meditron}} \\
          &       & \multicolumn{1}{l}{Taiyi,~\cite{luo2023taiyi}} \\
          &       & \multicolumn{1}{l}{BioMedGPT,~\cite{luo2023biomedgpt}} \\
         \cmidrule{2-3} 
          & \multirow{13}[0]{*}{Data} & \multicolumn{1}{l}{CheXpert,~\cite{irvin2019chexpert}} \\
          &       & \multicolumn{1}{l}{ROCO,~\cite{pelka2018radiology}} \\
          &       & \multicolumn{1}{l}{PMC-OA,~\cite{lin2023pmc}} \\
          &       & \multicolumn{1}{l}{VQA-RAD,~\cite{lau2018dataset}} \\
          &       & \multicolumn{1}{l}{RadVisDial,~\cite{johnson2019mimic}} \\
          &       & \multicolumn{1}{l}{K-PathVQA,~\cite{naseem2023k}} \\
          &       & \multicolumn{1}{l}{SLAKE,~\cite{liu2021slake}} \\
          &       & \multicolumn{1}{l}{PathChatInstruct,~\cite{lu2023foundational}} \\
          &       & \multicolumn{1}{l}{SAT-DS,~\cite{zhao2023one}} \\
          &       & \multicolumn{1}{l}{Chi-Med-VL,~\cite{liu2021slake}} \\
          &       & \multicolumn{1}{l}{MedMD,~\cite{wu2023towards}} \\
          &       & \multicolumn{1}{l}{MedICaT,~\cite{subramanian2020medicat}} \\
          &       & \multicolumn{1}{l}{ADNI,~\cite{petersen2010alzheimer}} \\
    \bottomrule
    \end{tabular}%
  \label{tab:addlabel}%
\end{table}%

\subsubsection{Image-text Multimodal model components}

Image-text Multimodal components achieve comprehensive understanding of text sentences and videos through text feature extraction, image feature extraction, and the fusion of text and image features. In this process, text feature extraction involves extracting a sequence of tokens from a sentence using a text encoder, while image feature extraction entails extracting a sequence of visual features from a video using a video encoder. These two separate feature sequences are then fused through a image-text multimodal fusion module, projecting them into a shared embedding space to generate a comprehensive, cross-modal representation. This approach to fusing text and image features facilitates a more holistic comprehension of the associations between text and images, providing richer information for image-text multimodal tasks. Next, we will sequentially introduce the developmental history of these components.

\textbf{Image Feature Extraction}
The evolution of computer vision technology has significantly advanced image feature extraction techniques~\cite{simonyan2014very,szegedy2015going,he2016deep,dosovitskiy2020image,carion2020end,lu2023multiscale,wan2023low}. Convolutional Neural Networks (CNNs) are widely utilized for this purpose, often achieving state-of-the-art results. Initially, an image is input into a CNN model, which then generates a feature map through convolution and pooling operations. This feature map is subsequently flattened into a one-dimensional vector and combined with text features. Inspired by the success of transformer-based language models, visual models have also begun to leverage this structure, for instance, introduce the Transformer architecture to vision models, surpassing CNN-based models on the ImageNet dataset.

\textbf{Text Feature Extraction}
Extracting representative features from textual data is crucial in image-text multimodal models, enabling the model to learn and understand the text better. This process involves converting the input text into vector representations. Traditional methods like Recurrent Neural Networks (RNNs) face challenges such as gradient explosion and vanishing problems. To overcome these, architectures like Long Short-Term Memory (LSTM)~\cite{graves2012supervised} and Gated Recurrent Unit (GRU)~\cite{chung2014empirical} have been proposed, incorporating mechanisms like attention. Recently, transformer-based models have set new benchmarks in text feature extraction. Pre-trained language models such as GPT, BERT ~\cite{devlin2018bert}, and LLAMA ~\cite{touvron2023llama} exemplify the thriving development of powerful tools for understanding textual data.

\textbf{Feature fusion}
In the context of numerous image-text multimodal tasks, information about images and text needs to be integrated~\cite{zhu2020deformable,afridi2021multimodal}. The most common method is concatenation ~\cite{xu2023multimodal,zhang2024survey}, which preserves the dimensional information of two features, while the attention mechanism is also a popular fusion method by dynamically adjusting the weights between images and text. Bilinear pooling~\cite{nam2023survey} method can capture high order interaction information between two features, and has strong expression ability. At the same time, image-text multimodal encoders are increasingly used to learn joint representations of images and problems through shared encoders ~\cite{yang2023code,mai2023excavating,xu2023multimodal}. In addition, a considerable amount of fusion technologies has focused on the attention mechanism~\cite{guo2022attention,niu2021review}. Co-attention ~\cite{liu2021dual,chaudhari2021attentive,rahate2022multimodal} allows the model to simultaneously focus on inputs from different modalities, enabling it to better capture the correlated information between them. Cross-attention ~\cite{huang2019ccnet,ahn2023star,rajan2022cross} emphasizes the influence of representations generated in one modality on another, achieving cross-modal information exchange.
\subsection{Generic model}
In this section, our primary focus revolves around universal pre-trained models. Specifically, in section \ref{3.2.1}, we review the overall architecture of universal models, aiming for a comprehensive understanding of their design and operational principles. And then we discuss the datasets utilized in universal pre-training tasks in section \ref{3.2.2}, investigating the diverse sources of information that models learn during training. To illustrate the differences and commonalities between general model architectures and those specifically designed for biomedical applications, we present Table \ref{tab:addlabel}. The upper portion of Table \ref{tab:addlabel} introduces some representative works related to general model architectures.
\subsubsection{Generic Image-text multimodal model architecture}
\label{3.2.1}
Image-text multimodal pre-trained models are generally divided into single-stream structures and dual-stream structures ~\cite{zhang2020single,iki2021effect,bugliarello2021multimodal,wu2023multimodal}. Single-stream pre-trained models fuse different information through attention mechanisms ~\cite{aladem2020single,li2021broken,afridi2021multimodal,ruan2022survey,du2022survey,long2022vision,wang2023survey}. The approach involves aggregating features from different modalities and interacting with the input model, with the most basic merging method being direct concatenation. Direct concatenation refers to mapping extracted visual and text features to the same dimension and directly concatenating visual feature sequences with text. Additionally, certain special data types provide a mapping relationship from text words to visual regions, allowing visual features to be directly inserted into text as words. In contrast, dual-stream models first encode two modalities using different structures and then achieve cross-modal fusion through mutual attention mechanisms ~\cite{du2022survey,huang2022developing,tang2022survey,yu2022survey}. Due to the increased number of network layers in visual pre-trained models, the preprocessing required before image-text multimodal interaction is relatively less for text features that have not undergone deep-level processing. Moreover, the structure of dual-stream models is relatively flexible, allowing for a flexible decision on how to handle different modalities before interaction based on specific situations. We will provide a detailed introduction to the classic models Contrastive Language-Image Pretraining (CLIP), Vision-and-Language Transformer (ViLT), Bootstrapping Language-Image Pretraining (BLIP), and Large Language and Vision Assistant (LLaVA). There is an evolutionary lineage among them. The CLIP model adopts a contrastive learning approach to learn the correspondence between text and images without annotated data and possesses powerful image-text comprehension capabilities. ViLT improves on CLIP to make the computation of the original pixels as small as the text markup, and focuses most of the computation on modal interaction modeling to improve inference efficiency. BLIP, also building upon CLIP, enhances its ability to understand image regions. Additionally, LLaVA combines CLIP with large language models, further enhancing the fusion and understanding capabilities of image-text information.

\begin{figure}[ht]
    \centering
    \includegraphics[width=1\linewidth]{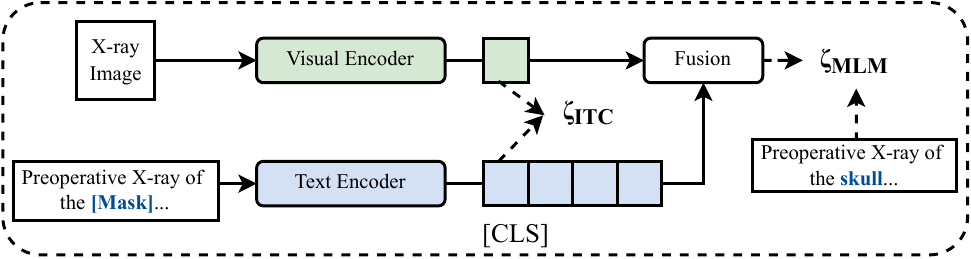}
    \caption{CLIP architecture(adapted~\cite{lin2023pmc}).}
    \label{fig:clip}
\end{figure}

CLIP (Contrastive Language-Image Pretraining), developed by OpenAI, represents an advanced dual-stream architecture in image-text multimodal pre-training models, utilizing the Vision-Language Transformer (ViT) framework. In this model, ViT acts as the image encoder, employing Transformer structures to effectively capture global information within images. As illustrated in Figure \ref{fig:clip}, CLIP distinguishes itself not only through its architecture but also through its innovative training methodology. It incorporates a text encoder that processes natural language text using a Transformer architecture similar to that of the image encoder. The unique aspect of CLIP lies in its use of contrastive learning as a training strategy. This approach aims to minimize the distance between the text and image features of the same sample in a shared embedding space, while maximizing the distance between features of different samples. Through this contrastive learning mechanism, CLIP is adept at learning and understanding the rich semantic relationships between text and images, setting it apart from other models.

\begin{figure}[ht]
    \centering
    \includegraphics[width=1\linewidth]{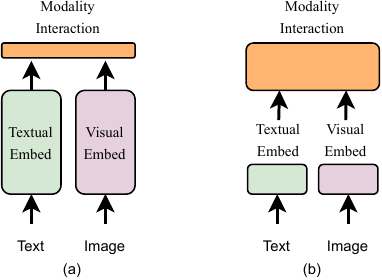}
    \caption{ViLT architecture(adapted~\cite{kim2021vilt}).}
    \label{fig:vilt}
\end{figure}

Some researchers have found that although current VLP improves performance on a variety of joint visual and language downstream tasks. These methods rely heavily on image feature extraction processes, most of which involve region supervision (such as object detection) and convolutional architectures. Therefore, there are problems in the inference efficiency and speed of the model. Vision-and-Language Transformer (ViLT) is thus proposed, which has a concise architecture as a VLP model with minimal visual embedding pipelines and follows a single-stream approach. The model initializes the interactive converter weights from a pre-trained ViT instead of BERT, and the processing of visual input is greatly simplified to the same convolution free way as for text input~\cite{kim2021vilt}. As shown in Figure\ref{fig:vilt}, the CLIP mentioned above belongs to Figure\ref{fig:vilt}(a), which uses a separate but equally expensive transformer embedder for each mode. Whereas, as shown in Figure\ref{fig:vilt}(b), the embedding layer of ViLT's raw pixels is shallow and computatively as small as text markup, the architecture focuses most of the computation on modal interaction modeling.

\begin{figure*}[ht]
    \centering
    \includegraphics[width=1\linewidth]{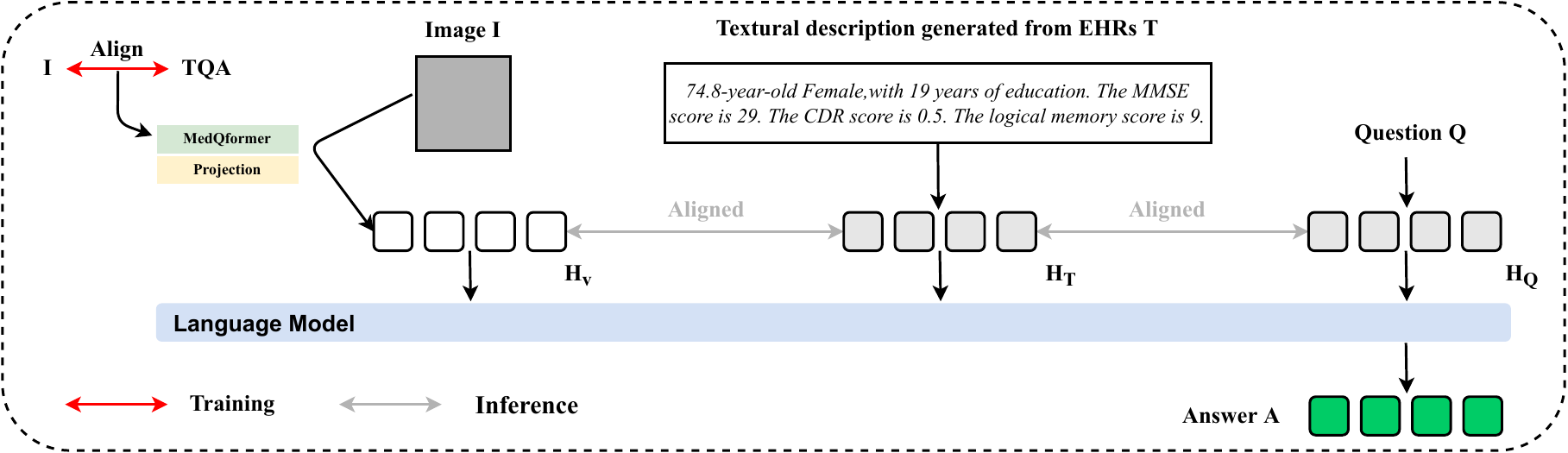}
    \caption{BLIP architecture(adapted~\cite{chen2023medblip}).}
    \label{fig:blip}
\end{figure*}

BLIP (Bootstrapping Language-Image Pretraining), introduced by Salesforce in 2022~\cite{Li_Li_Xiong_Hoi}, exemplifies the single-stream architecture within the realm of image-text multimodal frameworks, as depicted in Figure\ref{fig:blip}. This model innovatively combines understanding and generation tasks into a unified approach by leveraging cross-modal encoders and decoders, thereby achieving state-of-the-art performance in a variety of visual and language tasks. Central to BLIP's architecture is the image-text Multimodal Mixture of Encoder-Decoder (MED) structure. This comprehensive setup comprises two unimodal encoders—one for images and another for text—alongside an image-grounded text encoder and decoder. Such a configuration facilitates effective multitask pretraining and enhances BLIP's capability for transfer learning, making it a versatile tool in bridging visual and textual modalities.

\begin{figure}[ht]
    \centering
    \includegraphics[width=1\linewidth]{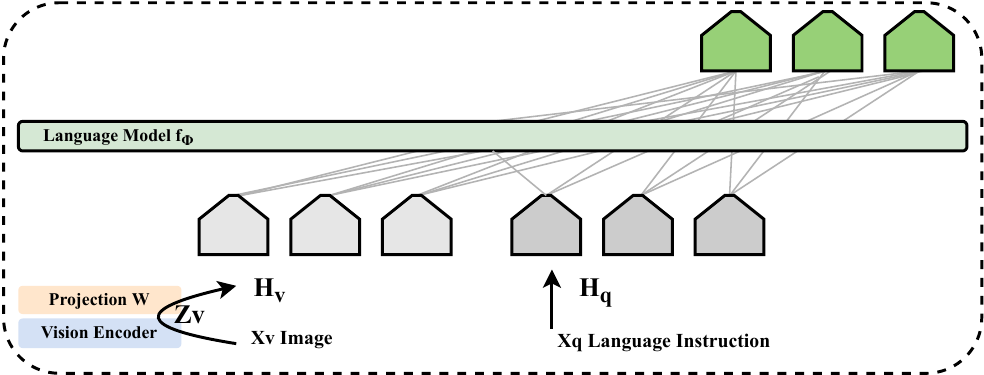}
    \caption{{LLaVA architecture(adapted~\cite{li2023llava}).}}
    \label{fig:llava}
\end{figure}

LLaVA represents an end-to-end trained, large-scale image-text multimodal model that integrates a visual encoder with a language model (LLM) to facilitate a universal understanding of both vision and language. Its primary objective is to effectively leverage the capabilities of pre-trained language models and visual models. As depicted in Figure \ref{fig:llava}, LLaVA exemplifies the single-stream architecture approach. The model selects Vicuna as its language model due to its exceptional performance in language tasks, as evidenced by publicly available checkpoints. For visual input \(X_v\), LLaVA utilizes the pre-trained CLIP visual encoder ViT-L/14, generating visual features \(Z_v = g(X_v)\). The model emphasizes grid features both before and after the last Transformer layer. Through a simple linear layer, LLaVA connects image features to the word embedding space. Specifically, it employs a trainable projection matrix \(W\) to convert \(Z_v\) into language embedding tokens \(H_v\), matching the dimensionality of the word embedding space in the language model. This efficient projection method facilitates rapid experimentation focused on data.

The exploration of image-text multimodal model architectures has led to the identification of two principal configurations: single-stream and dual-stream ~\cite{khan2022single,xu2020adversarial}. The single-stream approach integrates visual and textual data within a unified processing framework, optimizing the use of pre-trained components for enhanced efficiency. In contrast, the dual-stream configuration processes each modality separately before combining them, employing techniques such as contrastive learning and mutual attention mechanisms for a deeper image-text multimodal understanding. These methodologies underscore the field's varied strategies for the integration of visual and textual information, marking significant progress towards developing systems that achieve a comprehensive understanding of the complex interplay between vision and language.

\subsubsection{Generic Image-text Multimodal Task Data}
\label{3.2.2}
The success of image-text multimodal pretraining significantly depends on the effective utilization of diverse data resources. Image-text multimodal data, encompassing multiple sensory domains, is pivotal for models to acquire comprehensive knowledge, thereby enhancing their adaptability in real-world applications~\cite{benenson2022colouring, hudson2019gqa,kuznetsova2020open,wu2021fashion,changpinyo2021conceptual}.

Among the plethora of datasets, COCO (Common Objects in Context) ~\cite{lin2014microsoft} and Flickr30K~\cite{plummer2015flickr30k} are notably prominent. The COCO dataset offers a vast collection of image-text pairs, each annotated with details on object categories, relationships, and scene descriptions, making it invaluable for tasks requiring detailed visual understanding. Flickr30K, similarly, provides a rich set of image-text pairs, extensively used in image captioning and retrieval tasks.

The UNITER dataset~\cite{chen2020uniter} represents a significant effort in dataset amalgamation, compiling 5.6 million image-text pairs for comprehensive training. ImageBERT2~\cite{qi2020imagebert} expands upon this by assembling a dataset of 10 million web image-text pairs, aiming to further diversify the training material available for image-text multimodal models. Additionally, VL-BERT~\cite{su2019vl} enriches its training corpus with purely textual data from sources like BookCorpus and Wikipedia, alongside visual-textual pairs, to enhance language understanding capabilities.

To address the need for more diverse datasets in image-text multimodal research, recent efforts have introduced datasets such as Visual Genome~\cite{krishna2017visual}, which provides detailed annotations of images to facilitate tasks like visual question answering and relationship detection. The AI2D dataset~\cite{kembhavi2016diagram} focuses on diagram understanding, offering a unique challenge in interpreting graphical representations. Furthermore, the VQA dataset~\cite{antol2015vqa} and GQA~\cite{hudson2019gqa} have been instrumental in advancing visual question answering by providing questions and answers based on the content of images.

These datasets, with their extensive scale and rich annotations, serve as invaluable resources for researchers, driving the continuous enhancement of image-text multimodal model performance. A thorough exploration and application of these datasets are crucial for understanding the training dynamics and effectiveness of image-text multimodal models in capturing complex relationships between visual and textual information.

\subsection{Generic Model for Biomedicine}
This section is dedicated to discussing specialized pre-trained models and their applications within the biomedical domain. In section \ref{3.3.1}, we will thoroughly examine the architecture of image-text multimodal models tailored for the biomedical field, highlighting their pivotal role in synthesizing biomedical information. We aim to elucidate how these specialized models leverage universal pre-training techniques to tackle specific biomedical scenarios, thereby offering comprehensive support for biomedical analysis and decision-making. In section \ref{3.3.2}, our focus shifts to the challenges associated with biomedical data, including issues related to limited data volume and the narrow scope of problem categories. Our discussion aims to underscore strategies ensuring the effective deployment of these models in biomedical tasks, optimizing both their training processes and performance outcomes. Building on this discussion, the lower half of Table \ref{tab:addlabel} presents a selection of representative works in the field of biomedicine. This facilitates a deeper understanding of how specialized pre-trained models are transforming the biomedical domain.
\subsubsection{Image-text Multimodal Model Architecture for biomedicine}
\label{3.3.1}
An image-text multimodal biomedical model is an advanced framework that integrates diverse information sources in the biomedical field. By simultaneously processing and fusing information from various modalities such as text, imaging, laboratory data, and more, it offers comprehensive and in-depth support for biomedical research, diagnosis, and treatment. What sets this model apart is its unique ability to establish effective correlations between different data types, thereby providing a more comprehensive set of biomedical information, contributing to more accurate biomedical decision-making. Reflecting on the generic image-text multimodal model architectures discussed in section \ref{3.2.1}, the biomedical domain similarly benefits from specialized applications of these models. The following examples illustrate how image-text multimodal pre-training approaches are adapted to meet the unique challenges of biomedical scenarios, enhancing the field's capacity for comprehensive and in-depth biomedical support.

PMC-CLIP~\cite{lin2023pmc} is an example of applying the CLIP image-text multimodal model in the biomedical domain. PMC-CLIP relies on a pre-trained ResNet-50 for extracting image features, incorporating a trainable projection module. To align with biomedical outputs, a trainable projection module is added on top of ResNet-50 to bridge the gap between pre-trained visual and language embeddings. PMC-CLIP utilizes two projection variants. The first variant is based on MLP, employing a two-layer multilayer perceptron (MLP), while the second variant is based on a transformer, using a 12-layer transformer decoder with multiple learnable vectors as query inputs for the language encoder. To construct the output instruction, PMC-CLIP formulates the text as "Question: q, Answer:", encoding it using the language encoder. PMC-CLIP can also be applied to multiple-choice tasks by providing options and training it to output the correct choice as "A/B/C/D". The multi-modal decoder of PMC-CLIP is initialized with a 4-layer Transformer structure. Furthermore, PMC-CLIP transforms the generation task into a masked language modeling task, learning to generate predictions for masked tokens through the decoder module.

MedBLIP is a model designed for the integration of 2D visual encoding on 3D biomedical images, with a distinctive focus on reducing data requirements through the utilization of large pre-existing models. The core architecture of MedBLIP involves a query encoder that aligns visual features and leverages the capabilities of established large models. To bridge the dimensional gap between the 2D visual encoder and 3D biomedical scans, as well as align image features with textual ones, MedBLIP introduces a query network based on a transformer encoder. This network plays a crucial role in mapping visual features to a visual prefix in the language model (LM) decoder. Furthermore, MedBLIP incorporates a lightweight projection mechanism to adapt 3D image features, ensuring compatibility with the model's overall architecture. In terms of prompt structure, MedBLIP adopts an approach influenced by prior works such as ~\cite{li2022blip} and ~\cite{koh2023grounding}. Notably, MedBLIP introduces labeled descriptive strings between question and answer tokens. A distinctive choice made by MedBLIP is the placement of embeddings for both image and text descriptions before the question tokens. This prompt structure facilitates effective communication between the image and text modalities, enhancing the model's ability to comprehend and generate relevant responses in biomedical contexts.

LLaVA-Med is a model that shares similarities with the concept of prefix-tuning in language models (LMs), as presented in the work of Van Steenkiste et al.~\cite{van2023open}. The essence of LLaVA-Med lies in the integration of a trainable module between a frozen image encoder and a causal language model. In the referenced work by Van Steenkiste et al. ~\cite{van2023open}, a three-layer Multilayer Perceptron (MLP) network is employed to map visual features into a visual prefix. This approach is used in conjunction with pre-trained language models such as GPT2-XL ~\cite{radford2019language}, BioMedLM ~\cite{venigalla2022biomedlm}, and BioGPT ~\cite{mirmohammad2021conventional}. In contrast, LLaVA-Med adopts a different strategy by utilizing linear projection and a 7B LM ~\cite{van2023open,chiang2023vicuna}. The key innovation introduced by LLaVA-Med is its novel data generation method. This method leverages GPT-4 to extract freely available biomedical image-text pairs. By doing so, LLaVA-Med contributes to the field by expanding the repertoire of available data for training, thereby potentially enhancing the model's performance in biomedical applications. This novel data generation method sets LLaVA-Med apart from previous approaches and underscores its potential impact on advancing the capabilities of language models in the biomedical domain.

\subsubsection{Image-text Multimodal Task Data for Biomedicine}
\label{3.3.2}
Based on the biomedical multimodal model architecture discussed in Section \ref{3.3.1}, we will elaborate on the commonly used datasets, including the current issues faced by medical datasets, and the typical datasets utilized in this domain. 
Biomedical datasets often face limitations due to restricted data sources and the high level of expertise required for annotations, resulting in relatively small dataset sizes ~\cite{shin2018medical,tajbakhsh2020embracing,willemink2020preparing,garcea2023data}. Additionally, there is a lack of diversity in question categories within most existing biomedical datasets. Furthermore, each dataset is typically designed with specific objectives in mind, tailored to the dataset's focus and intended applications. While general datasets have aimed to address particular issues, such as answer balance, biomedical datasets are predominantly in the phase of topic and application scenario expansion ~\cite{seoni2023application,aminizadeh2023applications}. The majority of biomedical datasets contain fewer than 5,000 images, which presents challenges for training robust image representations. Proposed solutions to these challenges include utilizing alternative pre-trained models, leveraging additional datasets, employing contrastive learning, engaging in multi-task pre-training, and exploring meta-learning strategies. For instance, the LIST team utilizes an image encoder pre-trained on the CheXpert~\cite{irvin2019chexpert} dataset. The MMBERT team employs an auxiliary dataset named ROCO (comprising images and captions) ~\cite{DBLP:conf/miccai/PelkaKRNF18} for pre-training using a masked token strategy. The CPRD team introduces contrastive learning techniques, opting for pre-training with unlabeled images in a self-supervised scheme~\cite{herrett2015data}. The advantage of self-supervised training lies in its independence from image labels, which are often costly to acquire for biomedical images. Meanwhile, the MTPT-CSMA team leverages an auxiliary dataset from segmentation tasks.

In addition, an increasing number of biomedical datasets have been proposed in recent years, promoting the research of biomedical models.
PMC-LLaMA~\cite{wu2023pmc} compiles 30K textbooks from diverse sources, including open libraries, university libraries, and reputable publishers, encompassing a broad spectrum of biomedical disciplines. The preprocessing phase involves extracting text content from book PDFs and performing data cleaning to remove duplicates and filter content. Specifically, PMC-LLaMA discards irrelevant elements such as URLs, author lists, redundant information, document structures, references, and citations. Following this meticulous cleaning process, approximately 4 billion tokens are retained. These corpora encapsulate varied types of biomedical knowledge~\cite{wu2023pmc} and is primarily used for training the Visual Question Answering (VQA) task. The production process of LLaVA-Med training data was inspired by the PMC-LLaMA dataset, using open-source LLM as a pretraining model and constructing biomedical related instructions.
PMC-OA~\cite{lin2023pmc} is employed for training the task of image-text matching and is an important resource for training the PMC-CLIP model. It constitutes a comprehensive biomedical dataset that includes 2.4 million papers, which includes images of ear diseases, COVID-19 and fetal hypoxia, and corresponding diagnostic procedures such as X-ray, MRI, fMRI and CT. To preserve the diversity and complexity of PMC-VQA, PMC-OA utilizes 381K image-caption pairs derived from biomedical figures~\cite{lin2023pmc}.
MedICaT~\cite{subramanian2020medicat} is a multimodal dataset of images and texts from open access biomedical papers, which is used to evaluate the performance of the PMC-CLIP model. The estimated image types in the dataset mainly include radiology images (72\%), histology images (13\%), scope procedures (3\%), and other types of biomedical images (7\%). It has three data formats: image captions, subgraph caption annotations, and inline references within text. The MedICaT dataset aims to address the challenges of biomedical image retrieval and image to text alignment. Compared to the ROCO dataset, which only has images and captions, MedICaT also provides citation information within images and texts, which helps improve the model's ability to understand and retrieve images.
SLAKE~\cite{liu2021slake} is a biomedical dataset that contains rich visual and textual annotations, which is mainly utilized for training the VQA task. It is one of LLaVA-Med training data sources. It selects radiology images from three open source datasets~\cite{kavur2021chaos, wang2017chestx, simpson2019large}, including healthy and unhealthy cases. Experienced doctors use ITK-SNAP~\cite{yushkevich2006user} to perform detailed organ and disease labeling. SLAKE covers multiple body organs and diseases, and is composed of image types such as CT, X-ray, and MRI. The performance of models trained using the SLAKE dataset has improved in handling visual and knowledge-based problems, driving further development in this field.
ADNI Petersen et al. [2010]\cite{petersen2010alzheimer} are one of the main training and testing data for MedBLIP, used to improve its image-text matching ability. It contains imaging data corresponding to the brain structure, and its data type is MRI, PET, This dataset was obtained from 819 subjects (229 with normal cognition, 398 with MCI, and 192 with AD) who were followed up for 12 months. It is mainly used to study Alzheimer's Disease and is currently one of the authoritative datasets for studying Alzheimer's Disease.

In summary, this section delves into the critical components and challenges associated with image-text multimodal models across three specific domain tasks. Through the analysis of these tasks, we uncover the application potential of image-text multimodal models in various fields and the technical hurdles that must be surmounted. Concurrently, we provide an overview of the generic architecture, essential components, and requisite data for image-text multimodal models. From a general model overview to task-specific analyses, and real-domain application cases, this chapter offers profound insights into the workings of image-text multimodal text-based models and their practical applications. By using the biomedical domain as an exemplar, we demonstrate the applicability of image-text multimodal models.

\section{Challenges and future directions for multimodal modeling in image-text tasks}
\label{section4}
The application of general image-text multimodal models in the biomedical domain presents a series of challenges. These challenges need to be carefully considered and resolved to advance the development and practical application of this field. This section categorizes these challenges into external and internal factors, discusses them, and proposes solutions.

\subsection{External Factor}

In this section, we will introduce two main challenges from the perspective of external factors: multimodal data and computational pressure.

\textbf{Challenges for Multimodal Dataset}:
The development and refinement of large multimodal language models, particularly for tasks involving the integration of image and text data, encounter several pivotal challenges. These challenges are magnified in domains requiring high precision and reliability, such as biomedical research, yet they also pervade more general applications ~\cite{singh20203d,li2023medical}. This section outlines the key obstacles in curating multimodal datasets and suggests potential avenues for overcoming these hurdles.

\begin{itemize}
    \item \textbf{Data Scarcity and Annotation Quality}: High-quality, accurately annotated datasets are foundational for training effective multimodal models ~\cite{baltruvsaitis2018multimodal,bayoudh2022survey}. In biomedical research, the specificity and accuracy of annotations are paramount, given the potential implications for patient care and research outcomes. However, the scarcity of such datasets, compounded by the high cost and time required for expert annotation, poses a significant barrier. For example, annotating radiological images to detect subtle signs of disease requires extensive expertise and time, making the process both costly and slow ~\cite{bi2019artificial,wu2023cdt}.
    
    \item \textbf{Annotation Consistency and Subjectivity}: Ensuring annotation consistency across large datasets is challenging, especially when subjective interpretations come into play ~\cite{miceli2020between,metallinou2013annotation}. Variability among annotators can lead to inconsistencies, affecting the model's ability to learn reliable patterns. This issue is not confined to the biomedical domain but extends to any application where nuanced understanding of image and text data is critical ~\cite{mathews2019explainable,schindelin2015imagej}.
    
    \item \textbf{Contamination and Bias}: The presence of contaminated data or inherent biases within datasets can significantly impair a model's generalizability and performance across different contexts ~\cite{paullada2021data}. In biomedical research, biases related to demographic factors can skew model predictions, while in general applications, cultural and contextual biases may emerge. Addressing these biases is crucial for developing fair and effective models ~\cite{xu2022algorithmic,jacoba2023bias}.
    
    \item \textbf{Privacy and Ethical Considerations}: The collection and use of data, especially personal or sensitive information, are subject to stringent privacy and ethical standards ~\cite{dhirani2023ethical}. In biomedical research, protecting patient confidentiality while utilizing data for model training requires careful navigation of these standards. Similarly, in general applications, ensuring data subjects' rights and adhering to legal frameworks are essential ~\cite{javed2023ethical}.
\end{itemize}

To mitigate these challenges, several strategies can be employed:
\begin{itemize}
    \item Diversifying data sources and employing advanced bias detection tools can help create more balanced and representative datasets ~\cite{shahbazi2023representation,chu2023age}.
    \item Leveraging semi-supervised and weakly supervised learning techniques can reduce the reliance on extensive labeled datasets, making the annotation process more efficient ~\cite{maynord2023semi,ren2023weakly,li2024characterizing}.
    \item Implementing robust data privacy measures, such as encryption and differential privacy, ensures compliance with ethical standards and legal requirements ~\cite{dhirani2023ethical}.
\end{itemize}

Addressing the challenges associated with multimodal datasets is critical for advancing the capabilities of multimodal models in both biomedical research and broader applications. By adopting targeted strategies to overcome these obstacles, the field can move towards developing more accurate, fair, and efficient models.

\textbf{Computational Resource Demand}
The deployment of large pretrained models, particularly in computationally intensive fields like biomedicine, necessitates significant computational resources for both training and inference phases ~\cite{qiu2023large,wang2023accelerating}. This requirement often poses a challenge for researchers and limits the practical application of such models due to the need for high-performance computing devices, such as GPU clusters or cloud computing platforms. Moreover, the inference process of these models, especially when applied in real-time or at a large scale, demands considerable computational power.

To mitigate these challenges, it is imperative to explore and develop more efficient training algorithms and inference strategies. For instance:

\begin{itemize}
    \item \textbf{Efficient Training Algorithms:} Techniques such as pruning and knowledge distillation can significantly reduce the computational burden ~\cite{li2023model}. Pruning involves eliminating unnecessary weights from the model, while knowledge distillation transfers knowledge from a large model to a smaller, more efficient one.
    \item \textbf{Inference Strategies:} Implementing model quantization, which reduces the precision of the model's parameters, can decrease the computational resources required for inference without substantially compromising performance. Additionally, dynamic computation offloading to cloud services can offer a balance between computational demand and real-time processing needs ~\cite{li2023model,zhu2023survey}.
    \item \textbf{Leveraging Unlabeled Data:} Employing semi-supervised and weakly supervised learning techniques allows for the utilization of a small amount of labeled data alongside large volumes of unlabeled data. This approach reduces reliance on extensive labeled datasets, which are costly and time-consuming to produce ~\cite{ren2023weakly,taha2023semi}.
    \item \textbf{Model Compression:} Exploring model compression and quantization techniques can significantly decrease the model's size and computational complexity. This enables the deployment of pretrained models on devices with limited computational capacity, such as mobile phones or embedded systems ~\cite{msuya2023deep}.
\end{itemize}

These methods not only aim to reduce the computational resources required for training and deploying large models but also strive to maintain or even enhance the models' performance. For example, in the biomedical field, employing model compression techniques has enabled the deployment of complex diagnostic algorithms directly onto handheld devices, facilitating real-time patient monitoring and diagnosis without the need for constant connectivity to cloud computing resources. Such advancements underscore the potential of efficient computational strategies to broaden the accessibility and applicability of advanced machine learning models in resource-constrained environments.

\subsection{Intrinsic Factor}

First of all, we elaborate the unique challenges for image-text tasks. Then, the modal alignment issue will be discussed at the feature level.   Next, we will address the problem of catastrophic forgetting that occurs during training. And following this, we explore the challenge of interpretability during model inference.   Finally, we will examine the fairness concern that arises after the completion of model training.
  
\textbf{Unique Challenges for Image-Text Tasks}
The fusion of images and text within artificial intelligence applications, especially in the biomedical domain, introduces several unique challenges that require targeted solutions ~\cite{stahlschmidt2022multimodal,li2023artificial}. These challenges include:

\begin{itemize}
    \item \textbf{Semantic Alignment of Image and Text}: Precise semantic alignment between images and their corresponding textual descriptions is crucial ~\cite{yarom2024you}. For example, accurately linking an MRI image to its clinical report demands a deep understanding of visual patterns and complex biomedical terminologies ~\cite{azad2023foundational}.
    
    \item \textbf{Complex Multimodal Data Processing}: Integrating and interpreting visual and textual data simultaneously is inherently complex. A specific challenge is diagnosing from radiology images, where the model must correlate subtle visual cues with textual symptoms in patient records ~\cite{azad2023foundational}.
    
    \item \textbf{Cross-modal Representation and Matching}: Developing efficient cross-modal representation and matching methods is essential for tasks like automatic biomedical image captioning, requiring the generation of accurate textual descriptions for images ~\cite{zhao2020cross,sun2023scoping}.
    
    \item \textbf{Limited Perception Ability of LVLMs}: Current Large Vision Language Models (LVLMs) often struggle with incomplete or inaccurate visual information acquisition, such as distinguishing between benign and malignant tumors in histopathology images, which requires a high level of detail and precision ~\cite{liu2024survey}.
\end{itemize}

To overcome these challenges, the following strategies are proposed:

\begin{itemize}
    \item \textbf{Enhanced Semantic Alignment Techniques}: Implementing advanced deep learning architectures, such as attention mechanisms or transformers, can improve the modeling of relationships between visual and textual data. Applying these techniques to annotate biomedical images with diagnostic reports could significantly enhance semantic alignment accuracy ~\cite{choi2023transformer,he2023transformers,shamshad2023transformers}.
    
    \item \textbf{Sophisticated Cross-modal Learning Algorithms}: Algorithms that effectively learn and represent the complex relationships between images and text can facilitate the automatic generation of accurate descriptions for biomedical images, aiding in diagnostic processes ~\cite{bayoudh2023survey,karthikeyan2024novel}.
    
    \item \textbf{Incorporation of Domain Knowledge}: Integrating domain-specific knowledge, such as biomedical ontologies or databases, into the model's knowledge base can enhance its ability to accurately process and interpret multimodal data ~\cite{cai2023incorporating,murali2023towards}.
    
    \item \textbf{Model Compression and Optimization}: Exploring model compression techniques and optimizing algorithms for efficient processing can enable the deployment of LVLMs in resource-constrained environments, facilitating their use by healthcare professionals in remote areas ~\cite{sung2023ecoflap}.
\end{itemize}

Addressing these specific challenges with targeted strategies will unlock the full potential of multimodal models in image-text tasks within the biomedical field. This advancement will contribute to more accurate, efficient, and accessible biomedical diagnostics and research, pushing the boundaries of AI in healthcare.

\textbf{Multimodal Alignment and Co-learning}
The integration of visual and textual information, crucial in domains such as biomedical research, presents the challenge of multimodal alignment. This process is essential for establishing accurate correspondences between different modalities, such as images and their descriptive texts, within the same context ~\cite{jiang2021review}.

Challenges in multimodal alignment and co-learning include:

\begin{itemize}    
    \item \textbf{Complexity of Alignment Strategies}: Choosing the optimal alignment strategy from multiple viable options can be challenging, especially when dealing with long-term dependencies within the data. For instance, aligning sequential data from patient health records with corresponding sequential imaging studies involves understanding the temporal relationships and clinical relevance between textual and visual data ~\cite{wang2008aligning,shaik2023survey}.

    \item \textbf{Handling Noisy Inputs and Unreliable Labels}: In biomedical applications, data often come with noise and unreliable labels, complicating the learning process. Techniques to enhance the model's resilience, such as noise-robust learning algorithms and advanced label noise processing methods, are crucial for maintaining accuracy and reliability ~\cite{tao2019resilient,li2020label,nagarajan2024bayesian}.
\end{itemize}
To address these challenges, the following strategies are proposed:

\begin{itemize}    
    \item \textbf{Leveraging State-of-the-Art Deep Learning for Precise Similarity Measures}: The refinement of similarity measures between images and texts has benefited greatly from the latest deep learning advancements ~\cite{chen2023survey}. The introduction of models specifically designed for multimodal interactions, such as CLIP ~\cite{radford2021learning}, has provided new avenues for correlating visual and textual data. CLIP, by learning from a vast array of images and texts, excels at understanding images in the context of natural language descriptions, offering a nuanced approach to multimodal data analysis. This model, along with other advanced deep learning techniques, can be fine-tuned to achieve high alignment accuracy, effectively capturing the intricacies of multimodal data and enhancing cross-modal representation learning.
    
    \item \textbf{Implementing Advanced Sequence Modeling for Long-term Dependencies}: Recent advancements in sequence modeling, particularly with the advent of Transformer-based architectures, have shown significant promise in handling long-term dependencies in multimodal data ~\cite{tiezzi2024resurgence}. Models such as the Vision Transformer (ViT) ~\cite{han2022survey} for image processing and the latest iterations of Transformer-based language models (e.g., BERT ~\cite{devlin2018bert}, GPT~\cite{brown2020language}, and their successors ~\cite{mao2023gpteval}) are adept at capturing complex sequential relationships. These models leverage self-attention mechanisms to excel in understanding the sequential nature of multimodal data, thereby enhancing alignment and co-learning processes. 
    
    \item \textbf{Adopting Parallel and Non-parallel Learning Methods}: In collaborative learning scenarios, both parallel and non-parallel learning methods can facilitate effective knowledge transfer between modalities. This includes techniques like cross-modal attention mechanisms that allow the model to focus on relevant aspects of one modality based on the information from another ~\cite{song2024multi}.
\end{itemize}
By implementing these targeted strategies, the challenges of multimodal alignment and co-learning, especially in the biomedical research domain, can be effectively addressed. This will not only improve the accuracy and efficiency of multimodal models but also expand their applicability in critical areas such as biomedical diagnosis and treatment planning.

\textbf{Catastrophic Forgetting}
In the domain of artificial intelligence, particularly within biomedical research, the necessity for models to learn from and adapt to multiple tasks simultaneously introduces the challenge of catastrophic forgetting. This phenomenon occurs when the acquisition of new knowledge causes a model to overwrite or lose previously learned information, due to the limited capacity of the model's memory and the competitive nature of learning multiple tasks ~\cite{zhai2023investigating}.

Challenges related to catastrophic forgetting include:

\begin{itemize}
    \item \textbf{Retention of Task-Specific Knowledge}: Ensuring that a model retains essential information from each task without interference from subsequent learning experiences. For example, a model trained on diagnosing diseases from X-ray images might struggle to maintain its diagnostic accuracy if later trained on MRI images without strategies to prevent forgetting ~\cite{mellal2023cnn}.
    
    \item \textbf{Balancing New and Old Learning}: Finding an equilibrium where the model can continue learning new tasks while preserving the accuracy and relevance of previously learned tasks ~\cite{peng2021multiscale}. This balance is crucial in fields like biomedical research, where models may be required to process new types of biomedical data without losing proficiency in earlier tasks.
\end{itemize}

To effectively address catastrophic forgetting, several strategies have been proposed:

\begin{itemize}
    \item \textbf{Elastic Weight Consolidation (EWC)}: EWC mitigates forgetting by adding a regularization term to the loss function during training, which penalizes changes to weights that are important for previous tasks ~\cite{yin2021mitigating}. This approach allows the model to learn new tasks while retaining knowledge critical to past performances.
    
    \item \textbf{Progressive Neural Networks}: These networks combat forgetting by allocating separate network pathways for different tasks, allowing for the transfer of knowledge across tasks without interference. Each pathway can learn task-specific representations, while shared pathways facilitate knowledge transfer ~\cite{sharma2023advancing,chen2023progressive}.
    
    \item \textbf{Replay Mechanisms}: Implementing replay mechanisms involves periodically retraining the model on a subset of data from previous tasks, reinforcing the model's memory and preventing the decay of previously learned information. This technique is akin to human memory reinforcement through repetition ~\cite{chen2023our,zhou2023replay}.
    
    \item \textbf{Meta-Knowledge Storage and Retrieval}: Storing meta-knowledge, or knowledge about how to learn different tasks, enables a model to quickly adapt to new tasks using insights gained from past learning experiences. This strategy ensures that crucial information is cataloged and readily accessible for future task learning ~\cite{sen2021rdfm,xu2023unleashing}.
\end{itemize}

By incorporating these targeted strategies, models can be designed to learn continuously across a spectrum of tasks in biomedical research and beyond, without succumbing to the limitations imposed by catastrophic forgetting. This advancement not only enhances the model's universality and adaptability but also ensures that it remains effective and accurate across diverse applications.

\textbf{Model Interpretability and Transparency}
The complexity of large pretrained models, particularly those applied in biomedical research, poses significant challenges to interpretability and transparency. These models, while powerful, often operate as "black boxes," making it difficult to understand how they process and integrate multimodal data (images and text) to make decisions ~\cite{salahuddin2022transparency,joyce2023explainable}. This lack of clarity can hinder trust and limit the applicability of these models in sensitive areas such as biomedical diagnosis, where understanding the rationale behind a model's prediction is crucial.

Challenges in enhancing model interpretability and transparency include:

\begin{itemize}
    \item \textbf{Complex Internal Mechanisms}: The intricate architectures of large models make it challenging to trace how input data is transformed into outputs. For instance, in diagnosing diseases from biomedical images and patient notes, clinicians require insight into which features the model considers most indicative of a particular diagnosis ~\cite{huff2021interpretation}.
    
    \item \textbf{Limited Data Availability}: In specialized fields like biomedicine, the scarcity of available data can further complicate the task of building interpretable models. This scarcity makes it harder to validate the model's decision-making process against a broad spectrum of cases ~\cite{hoyos2023case}.
\end{itemize}

To address these challenges, several strategies can be adopted:

\begin{itemize}
    \item \textbf{Development of Interpretable Architectures}: Designing models with interpretability in mind from the outset, such as by incorporating modules that explicitly map the model's attention or reasoning processes ~\cite{hong2020human,hassija2024interpreting}. Techniques like attention mechanisms can highlight the parts of the input data most influential in the model's decision, providing insights into its internal workings.
    
    \item \textbf{Model Interpretation and Visualization Tools}: Utilizing tools and techniques that can dissect and display the model's decision pathways. For example, LRP (Layer-wise Relevance Propagation) ~\cite{montavon2019layer,achtibat2024attnlrp} or SHAP (SHapley Additive exPlanations) ~\cite{garcia2020shapley} values can be used to visualize the contribution of each input feature to the final decision, offering a clearer understanding of the model's rationale.
    
    \item \textbf{Incorporating Domain Knowledge}: Embedding domain-specific knowledge into the model, such as biomedical ontologies in biomedical applications ~\cite{smith2023biomedical}, can help guide the model's learning process and make its outputs more interpretable. This approach can also facilitate the validation of model predictions against established biomedical knowledge.
    
    \item \textbf{Indicators and Evaluation Techniques for Interpretability}: Developing metrics and evaluation frameworks that can quantitatively assess a model's interpretability and transparency ~\cite{nauta2023anecdotal}. This could involve creating benchmarks that measure the alignment of model predictions with human expert decisions or the clarity of the model's explanations.
\end{itemize}

By implementing these strategies, the field can make significant strides towards developing large pretrained models that are not only powerful but also interpretable and transparent. This advancement is particularly vital in biomedical research, where the ability to trust and understand model predictions can directly impact patient care and outcomes.

\textbf{Model Bias and Fairness Issues}
In the realm of artificial intelligence, particularly in biomedical research, the presence of inherent biases in large pretrained models can lead to unfair outcomes and potentially impact biomedical decisions and research findings ~\cite{baumgartner2023fair,chen2023algorithmic}. These biases may stem from skewed training data, biased model design, or both, necessitating a comprehensive approach to detect, understand, and mitigate them.

Challenges related to model bias and fairness include:

\begin{itemize}
    \item \textbf{Identification of Bias Sources}: Pinpointing the exact sources of bias within the data or model architecture. For example, a dataset predominantly composed of patient data from a specific demographic group may lead to models that perform poorly on data from other groups ~\cite{drukker2023toward}.
    
    \item \textbf{Development of Fairness Metrics}: Creating metrics that can quantitatively assess the fairness of model predictions across different groups ~\cite{jiang2023evaluating}. This is crucial in ensuring that models do not disproportionately favor or disadvantage any particular group, especially in clinical settings where such biases could affect patient care.
\end{itemize}

To effectively address these issues, several strategies can be adopted:

\begin{itemize}
    \item \textbf{Diverse and Representative Data Collection}: Ensuring that training datasets are diverse and representative of the population as a whole. This may involve actively seeking out underrepresented data or employing techniques like data augmentation to simulate diversity within the dataset ~\cite{singh2023unified}.
    
    \item \textbf{Bias Detection and Correction Techniques}: Utilizing advanced algorithms designed to detect biases in training data and model predictions. Once identified, techniques such as re-weighting training examples or adjusting model parameters can be applied to correct these biases ~\cite{prakhar2023bias}.
    
    \item \textbf{Fairness-aware Model Design}: Incorporating fairness considerations into the model design process itself ~\cite{shi2023towards}. This could include the use of fairness constraints or objectives during model training to actively promote equitable outcomes.
    
    \item \textbf{Continuous Monitoring and Evaluation}: Establishing processes for the ongoing monitoring of model performance and fairness post-deployment ~\cite{wang2023automated}. This ensures that models remain fair and unbiased as they encounter new data and are used in real-world applications.
\end{itemize}

By implementing these targeted strategies, the AI community can make significant strides towards developing models that are not only powerful but also fair and unbiased. This is particularly vital in biomedical research, where the stakes are high, and the impact of biases can have profound implications on patient outcomes and healthcare equity.

This section has provided a detailed examination of the challenges encountered in multimodal modeling for image-text tasks, particularly within the biomedical research field. It highlighted critical issues such as the complexities of multimodal dataset acquisition, computational demands, catastrophic forgetting, and the need for model interpretability, transparency, bias mitigation, and fairness. In response to these challenges, practical strategies and solutions were discussed, drawing upon recent advancements in machine learning and artificial intelligence. These strategies encompass the application of advanced algorithms for data processing, the exploration of innovative model architectures for enhanced learning, and the adoption of ethical practices for ensuring data diversity and model fairness. The insights offered aim to guide future research towards developing multimodal models that are not only technically proficient but also ethically sound and socially responsible ~\cite{lee2023multimodality}. The emphasis on continuous improvement and ethical considerations reflects the broader goal of leveraging AI to benefit biomedical research and other fields, while adhering to principles of responsible AI development ~\cite{brady2024developing}.


\section{Conclusion}
\label{section5}
This study explores the role of general image-text multimodal models and their technologies in advancing multimodal technologies in the biomedical field. From the perspective of general technology evolution, this paper analyzes the development trajectory of general image-text multimodal models and identifies the bottlenecks encountered when transferring to the biomedical field. Simultaneously, from the task perspective, it investigates the key components and technical challenges of image-text multimodal models in three specific domains: image captioning, image-to-text retrieval, and visual question answering (VQA). The paper outlines the general architecture, fundamental components, and essential data of image-text multimodal models. In the biomedical field, we provide a comprehensive overview from general model descriptions to the analysis of specific tasks and real-world application cases, thereby mapping out the application strategies of general models in biomedicine. Additionally, we summarize potential challenges and categorize them, proposing corresponding solutions. Looking forward, the research journey of image-text multimodal models remains long and fraught with challenges. With the continuous evolution of large language models and breakthroughs in multimodal technologies, we have reason to believe that these models will overcome existing limitations and unlock new possibilities. Therefore, we advocate for continued attention to the evolution of general technologies and their applications in the biomedical field, ensuring the advancement and effectiveness of model applications to meet practical needs.

\bibliographystyle{model1-num-names}
\bibliography{cas-refs}

\end{document}